\documentclass{article}

    \PassOptionsToPackage{numbers, compress}{natbib}
\usepackage[preprint]{neurips_2025}

\usepackage[utf8]{inputenc} % allow utf-8 input
\usepackage[T1]{fontenc}   % use 8-bit T1 fonts
\usepackage{url}                % simple URL typesetting
\usepackage{booktabs}     % professional-quality tables
\usepackage{amsfonts}     % blackboard math symbols
\usepackage{nicefrac}       % compact symbols for 1/2, etc.
\usepackage{microtype}   % microtypography
\usepackage{xcolor}         % colors
\usepackage{colortbl}
\usepackage{rotating}
\usepackage{wrapfig}
\usepackage{overpic}
\usepackage{caption}
\usepackage{wrapfig}
\definecolor{mygray}{gray}{0.93}

\RequirePackage{bbm}
\RequirePackage{array}
\RequirePackage{multirow}
\usepackage{bbding}
\usepackage{graphicx}
\usepackage{algorithm}
\usepackage{amsmath,amssymb}
\usepackage{enumitem}
\usepackage[noend]{algpseudocode}

\usepackage[normalem]{ulem}
\def\ie{\emph{i.e.}}
\def\eg{\emph{e.g.}}

\newcommand{\figref}[1]{Fig.~\ref{#1}}
\newcommand{\tabref}[1]{Tab.~\ref{#1}}

\newcommand{\secref}[1]{\S\ref{#1}}
\def\ourmodel{FreeVPS}
\def\ourdata{LU-VPS}

\definecolor{mygray2}{gray}{.92}
\definecolor{ourblue}{rgb}{0.657,0.657,1}
\definecolor{ourpink}{rgb}{1,0.8,0.8}

\usepackage[colorlinks=true, citecolor=ourblue]{hyperref} 

\title{\ourmodel: Repurposing Training-Free SAM2 for Generalizable Video Polyp Segmentation}

\author{%
Qiang Hu\textsuperscript{1}, 
Ying Zhou\textsuperscript{1},
Gepeng Ji\textsuperscript{2\textdagger},
Nick Barnes\textsuperscript{2},
Qiang Li\textsuperscript{1},
Zhiwei Wang\textsuperscript{1\textdagger}\\
\textsuperscript{1}Huazhong University of Science and Technology \quad
\textsuperscript{2}Australian National University \\  
\texttt{zwwang@hust.edu.cn} \\
\textsuperscript{\textdagger}Corresponding Author
}

\begin{document}

\maketitle

\begin{abstract}
% motivation
Existing video polyp segmentation (VPS) paradigms usually struggle to balance between spatiotemporal modeling and domain generalization, limiting their applicability in real clinical scenarios.
To embrace this challenge, we recast the VPS task as a \textit{track-by-detect paradigm} that leverages the spatial contexts captured by the image polyp segmentation (IPS) model while integrating the temporal modeling capabilities of segment anything model~2 (SAM2). However, during long-term polyp tracking in colonoscopy videos, SAM2 suffers from error accumulation, resulting in a snowball effect that compromises segmentation stability. We mitigate this issue by repurposing SAM2 as a video polyp segmenter with two training-free modules. In particular, the intra-association filtering module eliminates spatial inaccuracies originating from the detecting stage, reducing false positives. The inter-association refinement module adaptively updates the memory bank to prevent error propagation over time, enhancing temporal coherence.
Both modules work synergistically to stabilize SAM2, achieving cutting-edge performance in both in-domain and out-of-domain scenarios. Furthermore, we demonstrate the robust tracking capabilities of {\ourmodel} in long-untrimmed colonoscopy videos, underscoring its potential reliable clinical analysis.
\end{abstract}

\section{Introduction}
\label{sec:intro}

% what/why is colonoscopy
% Colorectal cancer (CRC) represents a major global health threat . 
% As the gold standard for detecting and treating CRC, colonoscopy utilizes a tiny endoscope to capture video of the digestive tract for polyp inspection, with polyp locations and edges playing a crucial role in both diagnosis and treatment. 
% To minimize misdiagnoses and mistreatments resulting from varying endoscopist skills, automatic polyp segmentation in colonoscopy videos is becoming increasingly vital for timely diagnosis and enhanced accuracy \cite{ji2024frontiers}.

% what is colonoscopy? >>> why AI is important for polyp identification? >>> introduce the background of this study (refer representative IPS and VPS studies)
Colonoscopy, the gold standard~\cite{shaukat2022current} for the detection of colorectal cancer, employs a mini endoscope to visualize the interior of the lower digestive tract, facilitating inspection and intervention pipelines. Accurate polyp identification is crucial during this procedure, yet endoscopists' differing experience can lead to misdiagnoses \cite{brenner2024reduction}. Recently, the emergence of intelligent analysis in colonoscopy \cite{ji2024frontiers} has gained wide attention, such as segmenting polyps from images \cite{fang2019selective, fan2020pranet, guo2022non, shi2023deep, chai2024querynet}/videos \cite{ji2022video, lu2024diff, yang2024vivim, hu2024sali}. These methods hold substantial promise in reducing diagnostic variability, improving polyp detection rates, and providing reliable assessments, thus enhancing procedural precision and patient outcomes.

% However, existing video polyp segmentation (VPS) methods~\cite{ji2022video,lu2024diff,hu2024sali} all adopt an end-to-end training framework, which is extremely resource-intensive, requiring large annotated videos. Although several VPS datasets~\cite{sanchez2020piccolo,ji2022video,ali2023multi} have been collected and made publicly available, they still exhibit weaknesses in data scale, lesion types, scenario variations, and differences in endoscopy vendors. Due to these limitations, current VPS methods struggle to effectively learn the comprehensive contextual features of polyps, hindering their ability to accurately identify complex and diverse polyps in real-world open scenarios.

% In contrast, collecting and annotating endoscopic images is much more efficient and cost-effective, allowing for broader coverage of diverse lesion types, digestive scenarios, and endoscopy vendors with fewer samples. This is one reason why image polyp segmentation (IPS) methods~\cite{fan2020pranet} have advanced more rapidly in recent years, benefiting from more comprehensive contextual distinctions between polyps and the background. However, since IPS methods cannot model temporal relationships, applying them directly to colonoscopy videos leads to temporally inconsistent results, limiting their applicability in temporal continuity assistance for endoscopists during polyp screening.

\begin{figure}[t]
\centering
\includegraphics[width=\textwidth]{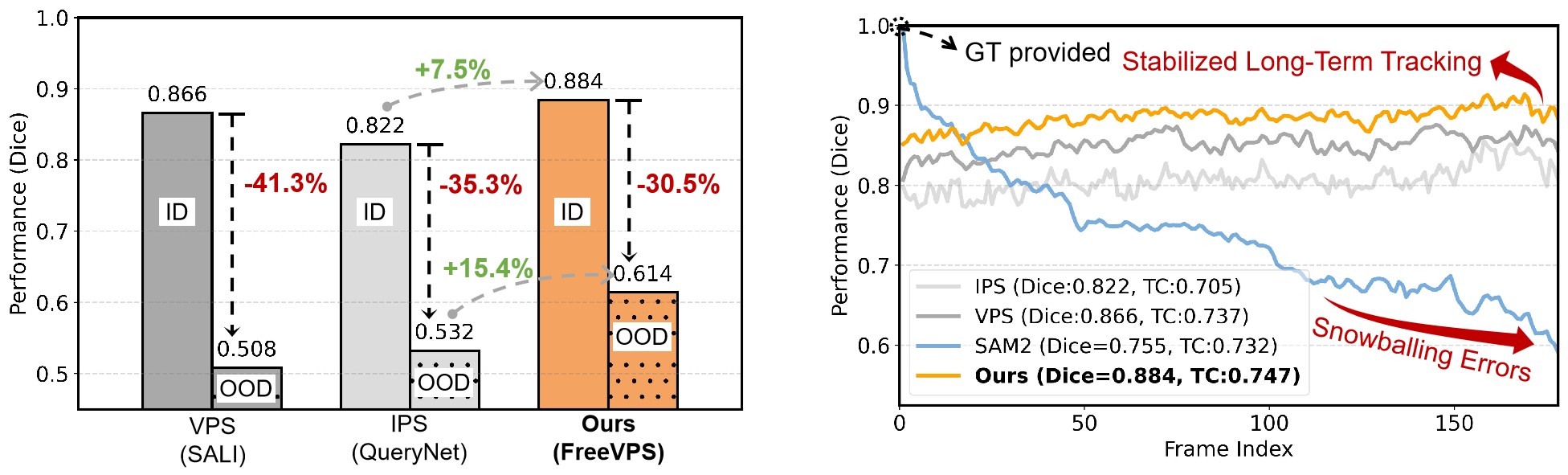}
% \vspace{-15pt}
\caption{
\textbf{\textit{(Left)} Cross-domain degradation.} 
VPS model (SALI \cite{hu2024sali}) outperforms IPS model (SALI \cite{hu2024sali}) in ID scenarios yet lags behind in OOD scenarios. By integrating QueryNet with the proposed \ourmodel, we boost both ID and OOD performance by $7.5\%$ and $15.4\%$, respectively, with less ID-to-OOD performance degradation.
% 
% Compared to IPS model (QueryNet~\cite{chai2024querynet}), VPS model (SALI \cite{hu2024sali}) exhibits better ID performance while lags behind in OOD scenarios. By empowering the QueryNet with the proposed \ourmodel, we improve its ID and OOD performance by $7.1\%$ and $7.5\%$ respectively, while with less ID-to-OOD degradation (only drop 30.5\%).
\textbf{\textit{(Right)} Snowball effect.} In temporal tracking, vanilla SAM2 \cite{ravi2024sam} exhibits a snowball effect, with errors accumulating and causing irreversible performance drops. In contrast, our \ourmodel~(orange line) maintains stable predictions over time, with highest temporal consistency score ($\text{TC}=0.747$).}
\label{fig:teaser_figure}
\end{figure}

% raise our core question; present the quantitative observation on existing methods
Despite these successes, we observe a negative correlation between spatiotemporal modeling and domain generalization in current approaches for video polyp segmentation (VPS). 
The left part of \figref{fig:teaser_figure} illustrates that the VPS model (SALI \cite{hu2024sali}) achieves a higher Dice score of $0.866$ in in-domain (ID) scenarios, surpassing the image polyp segmentation (IPS) model QueryNet \cite{chai2024querynet}, which scores $0.822$ in a per-frame inference manner. However, this advantage unexpectedly reverses under out-of-domain (OOD) conditions, where QueryNet overtakes SALI with better generalizability and experiences less performance degradation (IPS: $-35.3\%$ \textit{vs.} VPS: $-41.3\%$).
% 造成这种负相关现象的两大原因，一个是欠表达的空间上下文，另外一个就是优化偏差
% To investigate the underlying causes of this \textbf{cross-domain degradation} for video models, we identified two key factors. \textit{(a) Underrepresented spatial contexts.} Most VPS models \cite{ji2022video,lu2024diff,hu2024sali} are trained end-to-end on limited public video datasets \cite{sanchez2020piccolo,ji2022video,ali2023multi} due to constraints like annotation effort and budget. These datasets, compared to image-based datasets \cite{jha2020kvasir,wang2020improved,yue2023benchmarking}, are still less diverse in lesion types and scenario variations, resulting in an underrepresentation of spatial contexts crucial for OOD generalization. \textit{(b) Optimization bias.} Spatiotemporal modeling could bias video models' preference towards sequence-unique patterns \DJI{add a citation}, compromising their ability to effectively learn and generalize subtle spatial-contextual features, particularly under OOD conditions.
% 提出整文的核心问题，如何找到真实临床场景下的“时空建模”和“域泛化”之间的平衡点
% \NB{Given these factors,}
% \sout{In this end,} 
Given such a \textbf{cross-domain degradation} phenomenon, a natural question arises: \textit{``How can spatiotemporal modeling and domain generalization be effectively balanced in real clinical scenarios?''}

% what and why track-by-detect diagram
% , what can we benifit from this diagram for video polyp segmentation
To answer this question, we recast the VPS task into a \textit{track-by-detect} paradigm \cite{yang2019video,cao2020sipmask}, where polyps are first segmented independently in each frame and then associated across frames to produce temporally consistent masks. Importantly, in the \textit{detecting stage}, well-established foundations in the IPS community can be leveraged to obtain reliable predictions, either by utilizing off-the-shelf models or training a new model from scratch on various image datasets. The second \textit{tracking stage}, operating independently to strengthen spatiotemporal coherence of masks from stage one, can also function as a post-processor to refine existing models' predictions. 
% Therefore, we suggest such a paradigm shift to balance between spatiotemporal modeling and domain generalization in VPS task.
% 
However, considering current constraints on data scarcity, adopting the general-purpose track-by-detect models for VPS is non-trivial. Most methods \cite{cao2020sipmask,liu2021sg,yang2021crossover} rely heavily on extensive video data for robust representation, while this data-hungry behavior is costly and impractical in the medical field. Other modular methods \cite{veeramani2018deepsort,cheng2023tracking,du2023strongsort} integrate non-learning techniques (\eg, Kalman filter~\cite{kalman1960new}) and pre-trained models (\eg, re-identification \cite{luo2019strong}) for spatiotemporal association. However, they exhibit limited adaptability in colonoscopy scenarios, where the Kalman filter's linear assumption is incompatible with non-linear ego-motion of the colonoscope, and pre-trained re-identification models struggle to interpret medical visual patterns.
% 
% \DJI{revise}The observations indicate that the basic principle of our model design is to establish robust spatiotemporal associations with minimal dependence on video data.
% \DJI{revise this}In summary, the \textit{detection stage} operates independently with any IPS models, allowing us to focus on improving the spatiotemporal association ability for the \textit{tracking stage}.
Recall that the \textit{detection stage} operates independently of the IPS model, allowing us to concentrate on optimizing the \textit{tracking stage} in this study.

% is to spatially identify and segment polyps, which operate independently of any IPS model. In this study, we focus on improving the spatio-temporal association ability of \textit{tracking stage}. 

% sam2's superiority
Recently, SAM2 \cite{ravi2024sam} has achieved significant success in segmenting objects in videos, demonstrating robust tracking and generalization capabilities across general \cite{zhang2024segment} and medical \cite{ma2024segment} scenarios. However, as shown in the right part of \figref{fig:teaser_figure}, its chain-rule memory mechanism suffers from \textbf{snowballing errors} in colonoscopy videos, leading to cumulative performance degradation over time.
To alleviate such a snowball effect in the \textit{tracking stage}, we repurpose SAM2 as a zero-shot video polyp segmenter, named {\ourmodel}, which consists of two consecutive steps. 
% \DJI{more details about IAF and IAR; motivation is not smooth}
% The first step, we propose the intra-association filtering (IAF), which first spatially aligns multi-frame results from the IPS model to a unified reference frame in a time window, and then establishes the intra-association among the aligned results, effectively filtering out inconsistent results while retaining consistent results.
First, intra-association filtering (IAF) spatially aligns IPS's multiframe output to a unified reference frame within a time window. It then establishes intra-associations among these aligned results, effectively filtering out inconsistencies while preserving consistent predictions.
% The second step, we introduce inter-association refinement (IAR) to construct the inter-association between the IAF-prompted masks and SAM-propagated masks propagated from the prior window.
% Guided by this inter-association, we dynamically identify the appearance and disappearance of polyp instances throughout the video and optimize the segmentation results of consistent polyp instances.
Second, inter-association refinement (IAR) constructs inter-associations between IAF-prompted masks and SAM-propagated masks from the preceding time window. This step adaptively identifies the appearance and disappearance of polyp instances and refines the segmentation of consistent polyp instances across frames.
Our method synergistically integrates the IPS model with SAM2 in a training-free manner, achieving an improved
% optimal \NB{optimal means the best possible - do you just mean improved, or excellent or somthing like that?} 
balance between spatiotemporal modeling and domain generalization.  The experiments show that {\ourmodel} outperforms existing IPS and VPS methods on both ID and OOD VPS datasets. These findings highlight the clinical potential of {\ourmodel} for real-world colonoscopy analysis.
% \DJI{add core features within four lines}

In summary, our main contributions are as follows.
\textbf{(a)} To the best of our knowledge, we are the first to recast the VPS task into a \textit{track-by-detect} paradigm to allow each stage to specialize in distinct but synergistic sub-objectives. This paradigm effectively balances domain generalization in the \textit{detecting stage} and spatiotemporal modeling in  the \textit{tracking} stage.
\textbf{(b)} To alleviate the snowball effect of SAM2 in the \textit{tracking} stage, we repurpose it as a zero-shot video polyp segmenter that comprises two consecutive modules. The intra-association filtering (IAF) eliminates spatial inaccuracies originating from the \textit{detecting} stage, while the inter-association refinement (IAR) adaptively updates the memory bank to prevent error propagation over time.
\textbf{(c)} Our {\ourmodel} achieves cutting-edge performance on in-domain and out-of-domain generalizability, surpassing existing IPS and VPS paradigms. In a training-free manner, {\ourmodel} can boost the IPS model's ID performance by $7.5\%$ on SUN-SEG and OOD performance by $13.2\%$ and $15.4\%$ on PICCOLO and PolypGen, respectively. Importantly, on the untrimmed in-house video dataset, {\ourmodel} similarly significantly outperforms other methods.

\section{Related Works}
\label{sec:2}

\noindent\textbf{Image/Video Polyp Segmentation.}
With the advancement of deep learning, significant progress has been made in polyp segmentation, which can be primarily categorized into image- and video-based approaches. For image-based methods, seminal architectures like U-Net~\cite{ronneberger2015u} and PraNet~\cite{fan2020pranet} have established the technical foundation.
U-Net's encoder-decoder structure enables effective multi-scale feature fusion, while PraNet pioneered coarse-to-fine boundary refinement, excelling at segmenting ambiguous polyp boundaries.
Building upon these foundations, some methods have been further developed, such as introducing novel model architectures~\cite{dong2021polyp,xu2024polyp}, and unifying detection and segmentation~\cite{chai2024querynet}.
% innovations have focused on architectural modifications (e.g., transformer-based feature extractors) and integrated detection-segmentation paradigms. 
% model architectures like U-Net~\cite{ronneberger2015u} and PraNet~\cite{fan2020pranet} have laid critical foundations. U-Net's encoder-decoder structure enables effective multi-scale feature fusion, while PraNet pioneered coarse-to-fine boundary refinement, excelling at segmenting ambiguous polyp regions.
These methods are supported by diverse and comprehensive polyp image datasets~\cite{silva2014toward,bernal2015wm,tajbakhsh2015automated,vazquez2017benchmark,jha2020kvasir,yue2023benchmarking}, which capture varied polyp types and 
% \DJI{what is polyp modalities here?} \HQ{imaging modality: WLI or NBI}modalities, 
imaging modalities, enabling models to learn robust contextual representations. However, image-based frameworks inherently lack the ability to leverage temporal information, limiting their applicability to colonoscopy videos.
In contrast, video-based methods aim to exploit temporal dynamics for improved segmentation. Early works employed recurrent neural networks \cite{puyal2020endoscopic} or attention mechanisms \cite{ji2021progressively,ji2022video,fang2024embedding} to fuse multi-frame features within fixed time windows. The latest development, SALI~\cite{hu2024sali}, introduced memory banks to aggregate long-term temporal cues, achieving SOTA performance. Despite architectural progress, VPS remains constrained by data diversity and scale. For instance, the largest VPS dataset, SUN-SEG \cite{ji2022video}, contains only about $100$ distinct polyps across thousands of frames, reflecting the high cost of per-frame pixel-level annotation. Data scarcity limits video models' application to real-world clinical scenarios, where polyp appearance and motion vary significantly.

\noindent\textbf{Universal Temporal Propagation.}
% \DJI{to enrich this part and discuss more relevant studies}\HQ{finish}
In general scenarios, this category of methods is commonly referred to as semi-supervised video object segmentation, which propagates the initial ground truth (GT) segmentation of the first frame to the remaining entire video frames.
Perhaps most notably among the many solutions, \cite{oh2019video} innovatively introduced an external memory to explicitly store previously computed segmentation information, allowing the comprehensive use of past long-term segmentation cues.
Based on this, many of the later excellent models \cite{bhat2020learning,robinson2020learning,cheng2021rethinking,seong2022video,zhang2023boosting}, as well as the latest SAM2 \cite{ravi2024sam}, have adopted memory-based architectures.
Using the largest video dataset SA-V, SAM2 shows strong zero-shot generalization capabilities, enabling it to segment various videos. To further enhance the temporal stability of SAM2, some methods \cite{yang2024samurai,ding2024sam2long,videnovic2024distractor} optimized for memory storage and deletion, but have not focused on snowballing errors.
% 
% \noindent\textbf{\textit{Remarks.}} 
In contrast, our {\ourmodel} implements a synergistic \textit{track-by-detect} framework specially designed for the colonoscopy domain, where human-annotated videos are expensive and scarce. Specifically, a polyp image segmenter with specific semantic knowledge continuously guides SAM2, effectively suppressing error accumulation in the memory bank. Another important difference is that we adapt SAM2 to be an automatic object-specific video segmenter, whereas all of the above methods require human intervention.

% \DJI{you should focus on discussing everything related to `temporal propagation', but not dataset introduction.}\HQ{finish}\sout{Moreover, SAM2 construct the currently largest video segmentation dataset, called Segment Anything Video (SA-V) dataset, consisting of $35.5$M masks  across $50.9$K videos.}
% Similarly, the recently released SAM 2 also adopts this design, utilizing a memory bank to optimize long-range temporal modeling and ensure temporal segmentation consistency.
% Furthermore, it introduces a large-scale video dataset, from which the model learns powerful generalizable feature embeddings and temporal modeling capabilities.}
% Benefiting from these, SAM2 achieves SOTA performance and shows strong zero-shot generalization capabilities, enabling it to segment various videos.

\section{Methodology}
\label{sec:method}

% \subsection{Framework Architecture}
As in \secref{sec:3.1}, we first outline preliminaries of the introduced \textit{track-by-detect} paradigm -- IPS model for the \textit{detecting} stage and SAM2 for the \textit{tracking} stage -- which form the basis for our {\ourmodel}. Next, we introduce two key modules to construct \ourmodel: intra-association filtering (IAF in \secref{sec:3.2}) and inter-association refinement (IAR in \secref{sec:3.3}).

\subsection{Preliminaries}
\label{sec:3.1}

% \DJI{remove this paragraph? keep the necessary notation}\sout{
\noindent\textbf{IPS model.} 
% \DJI{add necessary notations related to IPS model} \HQ{finish}
% In this framework, the IPS model is responsible for actively perceiving the appearance of newly polyp instances, as well as continuously mitigating the \emph{error accumulation} of SAM2.
% Considering the model efficiency, we choose QueryNet~\cite{chai2024querynet}, a unified model for polyp detection and segmentation, as the IPS model.
This is used to obtain per-frame predictions during the \textit{detecting} stage, and it can be any well-trained model.
We denote its predictions as $\mathbf{M} = \{ m_i \}_{i=1}^{|\mathbf{M}|}$, where each element $m_i$ is a binary map indicating a single polyp instance, referred to as a \emph{segment}.
Notably, the different segments are non-overlapping: $m_i \cap m_j = \mathbf{0} (i \not= j)$, and $\mathbf{M}=\{ \mathbf{0} \}$ when no polyp is detected in the image.
% if no polyp is found in the image.

% Most existing IPS models are semantic segmenters, containing an encoder that encodes the image and a decoder that outputs the semantic segmentation result.
% Existing IPS models are mostly semantic segmenters that contain an encoder to encode the image and a decoder to output the semantic segmentation result.
% To facilitate subsequent instance-level association, we transform the semantic segmentation result into the instance segmentation result by simply defining each independent connected region as a polyp instance.
% Therefore, given an image, we denote the final output of the IPS model as $M = \{ m_i \}~(1 \leq i \leq |\mathcal{M}|)$, where the element $m_i$ is a binarized map that indicates each polyp instance, and we call it \emph{segment} in this work.
% Notably, the different segments do not overlap, \emph{i.e.}, $m_i \cap m_j = \mathbf{0} (i \not= j)$, and $M=\{ \mathbf{0} \}$ if there are no polyps in the image.

% Given an image, the output is a binarized semantic segmentation result that indicates the pixel-wise location of the polyp and background.
% However, to facilitate the subsequent construction of instance-level associations, we transform the semantic segmentation result into an instance segmentation result by defining each independent connected region as a polyp instance, 

\noindent\textbf{SAM2.}
% Driven by prompts (points, boxes) and/or stored memory, SAM2 can finely segment any object of interest in images or videos.
% In particular, benefiting from the large-scale video dataset SA-V~\cite{ravi2024sam} and the excellent memory-based architecture, SAM2 can overcome various video challenges such as motion blur, optical noise, and large object deformation to maintain long-term consistency of segmentation in videos.
% Benefiting from the memory bank that stores numerous past segmentation cues, SAM2 has remarkable long-term modeling capabilities.
We mainly let SAM2 perform its memory-based video tracking mechanism in the framework.
Specifically,  with video processing, several historical image-mask pairs $\{ \mathcal{I}_{mem}, \mathcal{M}_{mem} \}$ are stored in the memory bank.
% , where $\mathcal{I}_{mem}$ is a set of stored images, and $\mathcal{M}_{mem}=\{ \tau_n \}$ represents a set of mask trajectories.
% Each trajectory $\tau_n=\{ m_i^n \}$ is an ordered sequence of segmentation masks corresponding to a unique object identity, where $m_i^n$ denotes the binary mask of the $n$-th object on the $i$-th stored frame.
Given a query frame $\mathbf{I}_q$, SAM2 segments it by \emph{propagating} the stored masks based on the calculated \emph{correlation} between the stored images and the query frame:
\begin{equation}
\label{eq:1}
\mathbf{M}_{q}^{track} = \{ m_{q,n}^{track} \}_{n=1}^{N} = \operatorname{Prop} \big( \mathcal{M}_{mem},\operatorname{Cor}(\mathcal{I}_{mem}, \mathbf{I}_{q}) \big),
\end{equation}
where $\mathrm{Prop}(\cdot, \cdot)$ and $\mathrm{Cor}(\cdot, \cdot)$ represent the function of propagation and correlation in SAM2, respectively, $N$ indicates the total number of object trajectories stored in the memory bank, and $m_{q,n}^{track}$ denotes the tracking result corresponding to the $n$-th object.
% Technically, SAM2 is just a high-performance mask tracker that can only track single or multiple object instances already stored in its memory bank, but cannot actively discover new object instances.
Technically, SAM2 requires user-intervention and cannot actively discover new objects.
To this end, we merge the ability of the IPS model to actively segment polyps and the long-term spatial-temporal consistency of SAM2 in a framework, realizing a robust automatic VPS model.

\begin{figure}[t!]
\centering
\includegraphics[width=\textwidth]{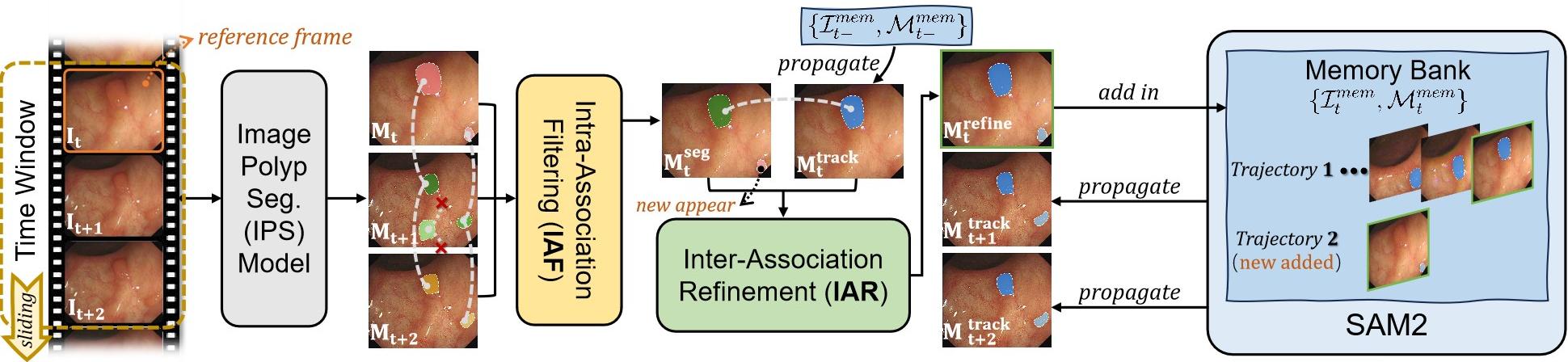}
\caption{Overview of \ourmodel. We adopt an $N(=3)$-length non-overlapping sliding window. Within each window, we first employ the Intra-Association Filtering (IAF) to discover temporally consistent predictions of the image polyp segmentation (IPS) model over multiple frames, Then, we establish the Inter-Association Refinement (IAR) module to refine the propagation results of SAM2 through consistent predictions of the IPS model. The refined results are finally added in the memory bank of SAM2, driving it to temporal-consistently segment the remaining frames in the window.}
\label{fig2}
\end{figure}

\subsection{Intra-Association Filtering for Consistency Discovery of the IPS Model}
\label{sec:3.2}
The rapid camera movements and unstable lighting conditions in endoscopic videos result in numerous discrete low-quality frames, and IPS models are prone to discrete false predictions due to their inability to exploit multi-frame features.
To generate more stable and reliable guidance from the IPS model for SAM2, we propose the IAF module.
% to associate the segmentation results of all frames within a time window, retaining consistent results while filtering out inconsistent results.

% To establish the instance-level association, by defining each independent connected region as a polyp instance, we transform the semantic segmentation results into the instance segmentation results, denoted as $ \{ \mathcal{M}_i = \{ m_j \}\}~(t \leq i \textless t+N, 1 \leq j \leq |\mathcal{M}_i|)$, where $m_j$ is a binarized map that indicates each polyp instance, and $\mathcal{M}_i$ includes a single zero map if the original semantic result is all zero.
% Note that we employ a semantic IPS model here as most state-of-the-art (SOTA) polyp segmentation methods are of this architecture, and the above transformation can be omitted if an instance IPS model is used.

\noindent\textbf{Cross-Frame Alignment.}
As shown in \figref{fig2}, given video frames $\{ \mathbf{I}_i \}_{i=t}^{t+T-1}$ withing a time window, where $t$ and $T$ denote the index of the first frame and the size of the window, respectively, the IAF module receives the corresponding segmentation result $\mathbf{\{M}_i \}_{i=t}^{t+T-1}$ generated by the IPS model.
% the IAF module receives the IPS model's segmentation result $\{ \mathbf{M}_i \}_{i=t}^{t+T-1}$ of the IPS model on frames $\{ \mathbf{I}_i \}_{i=t}^{t+T-1}$ within a time window, where $t$ and $T$ denote the index of the first frame and the size of the window, respectively.
To prevent frame variations from interfering with the subsequent spatial-coincidence-based association, the segmentation results of all frames should be aligned to a unified frame, called \emph{reference frame}.
Subsequently, we define the first frame $I_t$ in each time window as the reference frame and align the segmentation results of the remaining frames with it.
Considering that SAM2 is inherently well-suited for cross-image alignment, we utilize it to align the masks of all non-reference frame to the reference frame.
Specifically, we store the image-mask pair of each frame in turn in the memory bank and set the reference frame as the query frame to obtain the aligned mask of each frame:
% to the propagation results on the reference frame.
% By computing the correlation between $I_i$ and $I_t$, we propagate the segment to the first frame, resulting in the aligned segment $ \overline{\mathcal{M}}_i = \{ \overline{m}_j, 0 \leq j \textless |\overline{\mathcal{M}}_i| \}$, which is formulated as follows:
\begin{equation}
    \overline{\mathbf{M}}_i = \operatorname{Prop} \big( \mathbf{M}_i, \operatorname{Cor}(\mathbf{I}_i, \mathbf{I}_t) \big), t+1 \leq i \leq t+T-1,
\end{equation}
where $\overline{\mathbf{M}}_i$ is the spatially warped mask of ${\mathbf{M}}_i$ aligned to the reference frame.

\noindent\textbf{Tracklet Construction.}
To establish coherent mask tracklets, which record individual polyp instances consistently segmented by the IPS model across video frames, we propose a spatial coherence association mechanism.
Initially, we merge the reference frame's segmentation mask $\mathbf{M}_t$ with the aligned masks $\{\overline{\mathbf{M}}_i\}_{i=t+1}^{t+T-1}$ of the remaining frames, followed by sorting all segments to yield a new segment set:
% we group and sort the mask $\mathbf{M}_t$ of the reference frame with the aligned masks $\{\overline{\mathbf{M}}_i\}_{i=t+1}^{t+T-1}$ of the remaining frames into a segment set:
\begin{equation}
    \mathcal{S} = \{ s_k \} = \operatorname{Sort}({\mathbf{M}}_t \cup \left( \cup_{i=t+1}^{t+T-1} \overline{{\mathbf{M}}}_i \right) ),
\end{equation}
where $\mathrm{Sort}(\cdot)$ primarily orders segments by decreasing temporal proximity to the reference frame, and secondarily by decreasing area size.
% orders these segments in descending order of temporal proximity to the reference frame and descending order of area size.
Then, we compute a pairwise Intersection over Union (IoU) matrix $\mathbf{A} = [a_{ij}]_{|\mathcal{S}| \times |\mathcal{S}|}$, where the element $a_{ij}$ quantifies the IoU between segments $s_i$ and $s_j$.
By defining that $s_i$ and $s_j$ are successfully paired if $a_{ij}$ exceeds a predefined threshold $\theta$, we sequentially process the segments following their sorting priority in $\mathcal{S}$.
For each candidate segment $s_i$, if it is paired successfully with remaining $T-1$ segments from distinct frames, we create a set including this segment and its paired segments, called \emph{tracklet}.
Note that this formulation enforces the constraint that \emph{one segment is forbidden to be repeatedly included in more than one tracklet}.
In this way, assuming that we derive $P$ tracklets, denoted as $\{ tr_p \}_{p=1}^{P}$ and each tracklet $tr_p=\{s_k\}_{k \in \mathcal{K}_{p}}$, where $\mathcal{K}_p$ represents the index set of all segments belonging to the $p$-th tracklet in $\mathcal{S}$.
% We summarize the process of the above \emph{tracklet association} in Algorithm~\ref{alg:1}.
% \begin{figure}[htbp]
%     \begin{minipage}[t]{0.48\textwidth}
%         \small{
%         \begin{algorithm}[H]
%             \caption{Tracklet Association}\label{alg:1}
%             \begin{algorithmic}[1]
%                 \State x
%                 \State x
%             \end{algorithmic}
%         \end{algorithm}}
%     \end{minipage}
%     \hfill
%     \begin{minipage}[t]{0.48\textwidth}
%         \small
%         \begin{algorithm}[H]
%             \caption{Inter-Association Refinement}\label{alg:2}
%             \begin{algorithmic}[1]
%                 \State x
%                 \State x
%             \end{algorithmic}
%         \end{algorithm}
%     \end{minipage}
% \end{figure}

\noindent\textbf{Voting Filter.}
For each tracklet, we introduce a voting scheme to select the optimal segment that has the maximum sum of IoU scores with the rest segments in the tracklet, which is formulated as:
% and this selected segment serves as the final result on the reference frame of the polyp instance represented by this tracklet after the IAF module, which can be formulated as follows:
\begin{equation}
\label{eq:4}
    k^*_p = \underset{k \in \mathcal{K}_p}{\arg\max} \sum_{ l \in \mathcal{\mathcal{K}}_p, l \neq k} a_{kl},
\end{equation}
where $k^*_p$ indicates the index of the representative segment in the $p-$th tracklet.
We retain only these voted segments $\{s_{k^*_p}\}_{p=1}^{P}$ and serve them as the final output of the IAF module on the reference frame $\mathbf{I}_t$.
For clarity, we denote the result as $\mathbf{M}_t^{seg}=\{ m_{t,p}^{seg} \}_{p=1}^{P}$.

% We define that if $a_{ij}$ exceeds a threshold $\tau$, the \emph{i-th} and the \emph{j-th} segments are successfully paired.
% For $\mathcal{A}$, if there is the \emph{i-th} row that $\sum_{j=0}^{|\mathcal{S}|-1} \mathbbm{1}_\tau(a_{ij}) \geq T$, where $\mathbbm{1}_\tau(\cdot)$ denotes the binary function by threshold $\tau$, we consider the segment set $\{ s_j, a_{ij} \textgreater \tau \}$ to be \emph{consistent} throughout the time window.
% To obtain a unified representation of this set, we pick the indicator $j^*$ with the largest sum of paired IoU, which is defined as follows:
% \begin{equation}
%     j^* = {argmax}_j \sum_{a_{ik} \textgreater \tau} a_{jk}.
% \end{equation}
% By merging the unified representation of all sets, we obtain the final filtering result $ \mathcal{M}_{t}^{seg} = \{ m_{i}^{seg}, 0 \leq i \textless |\mathcal{M}_{t}^{seg}| \}$ of the image polyp segmentation model on $I_t$.
% Note that we guarantee that each segment in $\mathcal{S}$ will undergo only one candidate with the unified representation to avoid redundancy in the final result.

\subsection{Inter-Association Refinement for Enhancing SAM2}
\label{sec:3.3}
To empower SAM2 with the capability of dynamically detecting new polyp instances while mitigating the snowballing errors, we introduce the IAR module, converting SAM2 into a fully automatic video segmenter through IAF-guided operation.
% In order to equip SAM2 with the ability to actively discover the new polyps and alleviate its error accumulation problem,

\noindent\textbf{Inter-Association.}
% To empower SAM2 with the capability of dynamically detecting new polyp instances while mitigating error accumulation problems, we introduce the IAR module, which makes SAM2 well driven by the output of the IAF module.
Building upon the memory information preserved from the previous time window, denoted as $\mathcal{I}^{mem}_{t-}$, we let SAM2 track the reference frame of the current time window.
% First, we utilize the memory information retained at the end of the previous time window, denoted as $\{ \mathcal{I}^{mem}_{t-}, \mathcal{M}^{mem}_{t-}\}$, to drive SAM2 to track the reference frame of the current time window.
Assuming that $\mathcal{M}_{mem}^{-}$ stores $Q$ object trajectories, the process yields the tracking result ${\mathbf{M}}_t^{track}=\{ m_{t,q}^{track} \}_{q=1}^{Q}$.
Next, we perform the Hungarian algorithm to associate ${\mathbf{M}}_t^{track}$ and ${\mathbf{M}}_t^{seg}$.
Inspired by~\cite{carion2020end}, we pad the segment set with the smaller size using $\emptyset$ (zero map), ensuring the equal size  ($N= \text{max}(P,Q)$) for both sets.
Then we define the matching cost as IoU between segments and let the Hungarian algorithm \cite{kuhn1955hungarian} search for the optimal assignment of ${N}$ elements $\sigma \in \mathfrak{S}_{N}$ with the lowest cost:
\begin{equation}
    \hat{\sigma} = \underset{\sigma \in \mathfrak{S}_{N}}{\arg\min} \sum_{q}^{{N}} \\ - \operatorname{IoU}(m_{t,q}^{track}, m_{t,\sigma(q)}^{seg}).
\end{equation}
\noindent\textbf{Memory Refinement.}
To address the snowballing erros and perceive polyp appearance/disappearance in SAM2 tracking, we refine the memory by integrating ${\mathbf{M}}_t^{track}$ and ${\mathbf{M}}_t^{seg}$ according to the association result $\hat{\sigma}$.
The refinement is based on an operation, denoted as $\mathcal{O}(\cdot,\cdot |\hat{\sigma})$, which is conditionally defined as follows.
% Based on the association result, we perform an operation, denoted as $\mathcal{O}(\cdot,\cdot |\hat{\sigma})$, to integrate ${\mathbf{M}}_t^{track}$ and ${\mathbf{M}}_t^{seg}$, which is defined as follows.
\textbf{(a)} For the matched two segments, \emph{i.e.}, $m_{t,q}^{track}$ and $m_{t,\hat{\sigma}(q)}^{seg}$, we directly take their pixel-wise union $m_{t,q}^{track} \cup m_{t,\hat{\sigma}(q)}^{seg}$ as the final refined result on the reference frame of the polyp instance corresponding to the $q$-th trajectory stored in the memory. \textbf{(b)} For $m_{t,q}^{seg}$ matching with $\emptyset$, we consider it likely to be a newly appeared polyp instance founded by the IPS model. Furthermore, if the polyp appears in $\lambda_{1}$ consecutive time windows, we will keep $m_{t,q}^{seg}$ in the final refined result. \textbf{(c)} For $m_{t,p}^{track}$ matching with $\emptyset$, we consider that its corresponding polyp instance stored in the memory bank probably disappears. If it disappears in $\lambda_{2}$ consecutive time windows, the trajectory corresponding to the polyp instance will be removed from the memory.
% \begin{enumerate}[label=\arabic*),topsep=0.7pt,itemsep=2pt,leftmargin=20pt]
%     \item For the matched two segments, \emph{i.e.}, $m_{t,q}^{track}$ and $m_{t,\hat{\sigma}(q)}^{seg}$, we take their union $m_{t,q}^{track} \cup m_{t,\hat{\sigma}(q)}^{seg}$ as the final refined result on the reference frame of the polyp instance corresponding to the $q$-th trajectory stored in the memory.
%     \item For $m_{t,q}^{seg}$ matching with $\emptyset$, we consider it likely to be a newly appeared polyp instance founded by the IPS model, and if the polyp appears in $\lambda_{1}$ consecutive time windows, we will keep $m_{t,q}^{seg}$ in the final refined result.
%     \item For $m_{t,p}^{track}$ matching with $\emptyset$, we consider that its corresponding polyp instance probably disappears. If it disappears in $\lambda_{2}$ consecutive time windows, the trajectory corresponding to the polyp instance will be removed from the memory.
% \end{enumerate}
The final integrated result $\mathbf{M}_{t}^{refine}=\mathcal{O}(\mathbf{M}_{t}^{track}, \mathbf{M}_{t}^{seg}|\hat{\sigma})$ is regarded as the final segmentation result of the reference frame and added in the memory bank (as illustrated in \figref{fig2}).
In particular, newly detected polyp instances are assigned unique identities in memory.
The updated memory bank, denoted as $\{ \mathcal{I}^{mem}_t, \mathcal{M}^{mem}_{t}\}$, is utilized to track the remaining frames in the current time window using Eq.~(\ref{eq:1}), obtaining $\{ \mathbf{M}_i^{track} \}_{i=t+1}^{t+T-1}$.

\begin{table*}
\begin{center}
\caption{Comparisons on an ID dataset (SUN-SEG \cite{ji2022video}) and two OOD datasets (PICCOLO~\cite{sanchez2020piccolo}, PolypGen~\cite{ali2023multi}). ``$\dagger$" indicates that SAM2 is fully fine-tuned on video data $\mathcal{D}_V$ with officical code.
% , and this implementation is based on the official code provided by SAM2.
}
\vspace{-6pt}
\renewcommand\arraystretch{0.95}
\setlength{\tabcolsep}{2.7 pt}
% \resizebox{0.95 \linewidth}{!}{
\footnotesize
\begin{tabular}{lc|cccc|cccc|cccc}
\hline

\hline
\rowcolor{mygray}
&&\multicolumn{4}{c|}{SUN-SEG (\secref{sec:4.2})} &\multicolumn{4}{c|}{PICCOLO (\secref{sec:4.3})}& \multicolumn{4}{c}{PolypGen (\secref{sec:4.3})} \\
\rowcolor{mygray}
Models &Train Data &Dice &IoU &MAE &TC &Dice &IoU &MAE &TC &Dice &IoU &MAE &TC \\
\hline
\hline
\multicolumn{8}{l}{$\blacktriangleright$~\emph{Image Polyp Segmentation Models}}\\
% PraNet~\cite{fan2020pranet} &$74.4$ &$64.6$ &$4.2$ &$68.8$ &$61.7$ &$5.6$ &$65.4$ &$57.3$ &$5.0$ &$62.5$ &$54.2$ &$6.4$\\
PolypPVT~\cite{dong2021polyp} &$\mathcal{D}_V$ &$78.6$ &$70.6$ &$4.0$ &$69.4$ &$51.9$ &$42.7$ &$14.7$ &$64.2$ &$35.5$ &$28.2$ &$12.3$ &$60.0$\\
% Polyper~\cite{shao2024polyper}$ &$80.1$ &$71.2$ &$3.1$ &$73.5$ &$64.6$ &$4.4$ &$74.7$ &$65.8$ &$4.0$ &$71.1$ &$63.0$ &$4.1\\
Polyper~\cite{shao2024polyper} &$\mathcal{D}_V$ &$79.7$ &$71.1$ &$3.7$ &$69.8$ &$54.4$ &$43.3$ &$13.8$ &$64.2$ &$38.3$ &$32.7$ &$10.5$ &$61.2$\\
QueryNet~\cite{chai2024querynet} &$\mathcal{D}_V$ &$80.5$ &$72.1$ &$3.4$ &$70.1$ &$58.9$ &$51.3$ &$10.5$ &$64.5$ &$41.2$ &$34.3$ &$10.3$ &$61.0$\\
QueryNet~\cite{chai2024querynet} &$\mathcal{D}_V$+$\mathcal{D}_I$ &$82.2$ &$73.8$ &$2.8$ &$70.5$ &$68.0$ &$62.2$ &$7.7$ &$66.2$ &$53.2$ &$46.7$ &$4.7$ &$63.4$\\
% Polyper~\cite{shao2024polyper} &$90.1$ &$84.3$ &$1.4$ &$86.4$ &$79.5$ &$2.7$ &$79.9$ &$71.5$ &$3.4$ &$77.8$ &$69.6$ &$3.5$\\
% Polyper~\cite{shao2024polyper} &$90.8$ &$84.4$ &$1.6$ &$87.0$ &$80.7$ &$2.8$ &$82.1$ &$74.6$ &$3.0$ &$79.6$ &$72.3$ &$3.3$\\
\hline
\multicolumn{8}{l}{$\blacktriangleright$~\emph{Video Polyp Segmentation Models}}\\
% 2/3D~\cite{puyal2020endoscopic}$ &$85.6$ &$79.7$ &$2.0$ &$80.9$ &$73.5$ &$3.6$ &$72.2$ &$63.2$ &$4.6$ &$70.6$ &$61.5$ &$3.9\\
% PNS-Net~\cite{ji2021progressively} &$75.9$ &$68.7$ &$3.5$ &$-$ &$48.6$ &$38.4$ &$14.6$ &$-$ &$31.4$ &$26.2$ &$12.5$ &$-$\\
PNS+~\cite{ji2022video} &$\mathcal{D}_V$ &$80.9$ &$73.3$ &$3.4$ &$73.3$ &$53.1$ &$44.9$ &$14.2$ &$67.8$ &$44.1$ &$37.5$ &$8.1$ &$64.2$\\
AEN~\cite{fang2024embedding} &$\mathcal{D}_V$ &$81.9$ &$72.4$ &$3.4$ &$72.9$ &$58.2$ &$50.5$ &$11.3$ &$67.6$ &$44.8$ &$38.1$ &$8.1$ &$63.9$\\
Diff-VPS~\cite{lu2024diff} &$\mathcal{D}_V$ &$82.7$ &$75.3$ &$2.9$ &$72.8$ &$63.4$ &$55.0$ &$8.2$ &$67.8$ &$44.3$ &$37.2$ &$8.4$ &$64.3$\\
Vivim~\cite{yang2024vivim} &$\mathcal{D}_V$ &$83.2$ &$74.4$ &$3.0$ &$73.2$ &$66.2$ &$58.5$ &$8.3$ &$68.1$ &$48.4$ &$42.4$ &$6.6$ &$64.3$\\
SALI~\cite{hu2024sali} &$\mathcal{D}_V$ &$86.6$ &$80.0$ &$2.3$ &$73.7$ &$69.7$ &$62.7$ &$6.9$ &$68.4$ &$50.8$ &$45.0$ &$5.1$ &$64.5$\\
\hline
\multicolumn{8}{l}{$\blacktriangleright$~\emph{Image segmentor: QueryNet}}\\
% {\ourmodel}$ &$84.0$ &$75.7$ &$2.2$ &$79.3$ &$71.4$ &$3.8 $ &$77.9$ &$69.7$ &$3.7$ &$76.8$ &$67.8$ &$3.4\\
\textbf{+ {\ourmodel}} &$\mathcal{D}_V$ &$87.5$ &$80.3$ &$2.0$ &$74.2$ &$72.8$ &$66.4$ &$6.4$ &$69.2$ &$54.0$ &$48.6$ &$4.4$ &$65.3$\\
\textbf{+ {\ourmodel}} &$\mathcal{D}_V$+$\mathcal{D}_I$ &$88.4$ &$82.5$ &$1.6$ &$74.7$ &$77.0$ &$71.5$ &$6.0$ &$70.0$ &$61.4$ &$54.3$ &$3.3$ &$65.8$\\
\textbf{+ {\ourmodel}}$^\dagger$ &$\mathcal{D}_V$+$\mathcal{D}_I$ &$\mathbf{90.2}$ &$\mathbf{84.8}$ &$\mathbf{1.3}$ &$\mathbf{75.1}$ &$\mathbf{83.1}$ &$\mathbf{77.4}$ &$\mathbf{5.5}$ &$\mathbf{70.2}$ &$\mathbf{65.3}$ &$\mathbf{59.1}$ &$\mathbf{2.8}$ &$\mathbf{66.2}$\\
% {\ourmodel}$^\dag$ &\textbf{94.0}&\textbf{88.7} &\textbf{0.9} &\textbf{91.5} &\textbf{85.8} &\textbf{1.5} &\textbf{88.1} &\textbf{81.5} &\textbf{1.9} &\textbf{87.1} &\textbf{80.0} &\textbf{1.8}\\
\hline

\hline
\end{tabular}
% \end{small}
% }
\label{table:1}
\end{center}
\end{table*}

% \vspace{-20pt}
\section{Experiments}
\label{sec:exp}

\subsection{Experimental Protocols}

\noindent\textbf{Datasets.}
% For the video dataset, we choose SUN-SEG~\cite{ji2022video} as in-domain \DJI{in-domain} \HQ{finish}experimental dataset, its train set (denoted as $\mathcal{D}_V$) contains $112$ video clips, with a total of $19, 544$ frames, and its test set includes four sub-test sets, with a total of $173$ video clips and $29,592$ frames.
We adopt the widely-used VPS dataset, SUN-SEG \cite{ji2022video}, as the ID data source. Its training set, referred to as $\mathcal{D}_V$, comprises $112$ video clips totaling $19, 544$ frames, and its testing set contains $173$ video clips totaling $29,592$ frames.
% \DJI{you said split into four sub-sets, but you only provide a set of evaluation results on SUN-SEG. This might be confusing for reviewers}\HQ{finish}
% \sout{Additionally, the test set is partitioned into four sub-test sets}, which covers a total of $173$ video clips and $29,592$ frames.
% Seen-Easy ($33$ clips/$4, 719$ frames), Seen-Hard ($17$ clips/$3, 882$ frames), Unseen-Easy ($86$ clips/$12, 351$ frames), and Unseen-Hard ($37$ clips/$8, 640$ frames). Easy/Hard indicates that difficult levels to be segmented, and Seen/Unseen indicates whether the clips are sampled from the same video as the train set.
% In addition, we introduce two VPS datasets, PICCOLO~\cite{sanchez2020piccolo} and PolypGen~\cite{ali2023multi}, to test the OOD generalization.
% Importantly, to compensate for the lack of real clinical scenarios,\DJI{real clinical scenarios}\HQ{finish} we construct an in-house long-untrimmed VPS dataset, {\ourdata}, which contains a total of $9$ positive and $6$ negative videos, with video timing ranging from $20$ minutes to $45$ minutes at $15$ frames per second, and the box and mask annotations are provided by three professional endoscopists.
To quantify generalizability in OOD scenarios, we also include two public VPS datasets: 
% \DJI{provide data statistic for each}\HQ{finish} 
PICCOLO \cite{sanchez2020piccolo} ($39$ video clips$, 3,433$ frames) and PolypGen \cite{ali2023multi} ($23$ video clips, $8,037$ frames). However, these VPS benchmarks usually consist of human-trimmed videos, assuming each frame contains at least one polyp. This assumption restricts the model's applicability to real-world clinical scenarios, where frames without polyps or other findings are common.
% we construct a novel untrimmed VPS benchmark to test the performance of the model in real clinical practice.
To address this gap, we create an in-house dataset, {\ourdata}, comprising nine positive and six negative
% $9$\NB{nine} positive and $6$\NB{six} negative 
long untrimmed colonoscopy videos from real clinical procedures. 
% These videos range from $20\!\sim\!50$ minutes at $15$FPS. To ensure reliability, our {\ourdata} is annotated by three professional endoscopists with $15\!\sim\!20$ years of clinical experience, providing $59,104$ bounding boxes and $48,282$ segmentation masks for polyps. 
The video durations span $20\!\sim\!50$ minutes at $15$FPS. To guarantee reliability, three expert endoscopists with $15$-$20$ years of clinical experience annotated our {\ourdata}, yielding $59,104$ bounding boxes and $48,282$ masks for polyps. For more details like data statistics and visualizations, refer to the \textcolor{magenta}{\textsc{Appendix}}.
% \DJI{with how many years experience in this area?}\HQ{finish}, 
% \NB{You should think about describing this in more detail in the Supp. Domain specialists are likely to want additional detail of new datasets. See what has been detailed for other similar datasets. Perhaps refer to this here.}
% To explore potential gains from abundant image resources, we further introduce six public IPS data (\ie, ClinicDB~\cite{bernal2015wm}, Kvasir-SEG~\cite{jha2020kvasir}, ColonDB~\cite{tajbakhsh2015automated}, EndoScene~\cite{vazquez2017benchmark}, ETIS~\cite{silva2014toward}, and PS-NBI2K~\cite{yue2023benchmarking}) totaling $5,026$ images. These images (denoted as $\mathcal{D}_I$) will exclusively be used to train IPS model for experimental comparison.
We introduce six public IPS datasets to fully leverage abundant image resources: ClinicDB \cite{bernal2015wm}, Kvasir-SEG \cite{jha2020kvasir}, ColonDB \cite{tajbakhsh2015automated}, EndoScene \cite{vazquez2017benchmark}, ETIS \cite{silva2014toward}, and PS-NBI2K \cite{yue2023benchmarking}, totaling $5,026$ images. These images, termed $\mathcal{D}_I$, will be used exclusively to train the IPS model for robust spatial representation.

\noindent\textbf{Evaluation Metrics.} 
% \DJI{rewrite this paragraph} \sout{Depending on the clinical concern of whether the model can distinguish between polyps and surrounding tissues at the pixel level without the need for instance-level distinction among different polyps, and the type of data annotation, we mainly validate the performance of the model on semantic segmentation in this work, so we transform the outputs of all methods into semantic masks,}\HQ{finish}
We mainly focus on verifying the performance of the model in semantic video segmentation and employ four metrics to evaluate VPS performance, including Dice, intersection over union (IoU), mean absolution error (MAE) and temporal consistency (TC)~\cite{liu2020efficient}.

\noindent\textbf{Implementation Details.} 
% \DJI{double check out the revision, to ensure nothing ignored}\HQ{finish}
% \label{sec:4.2}
Recall that {\ourmodel} functions as a training-free post-processor to refine temporal masks from any IPS model. In subsequent experiments, we used the cutting-edge model, QueryNet \cite{chai2024querynet}, as a representative example to build \ourmodel. More validations on other IPS models are provided in the \textcolor{magenta}{\textsc{Appendix}}. {\ourmodel} is built upon SAM2 \cite{ravi2024sam} with Hiera-B+ \cite{ryali2023hiera}, using the PyTorch framework \cite{paszke2019pytorch} and is inferred on a single NVIDIA GeForce RTX 4090 GPU with $24$GB memory. Based on the ablation studies in \secref{sec:diagnostic_experiments}, we set the length of time windows $T=3$ and the paired threshold $\theta=0.5$. Moreover, we empirically set $\{\lambda_1,\lambda_2\}=\{1,3\}$ in memory management.

\noindent\textbf{Baseline Models.} As shown in \tabref{table:1}, we select several of the latest polyp segmentation models, including three IPS models \cite{dong2021polyp,shao2024polyper,chai2024querynet} and five VPS models \cite{ji2022video,fang2024embedding,lu2024diff,yang2024vivim,hu2024sali}. Notably, current VPS models are all trained in the SUN-SEG training set $\mathcal{D}_V$. Therefore, we retrain three IPS models on the same set $\mathcal{D}_V$ to ensure a fair comparison between image- and video-based models.

% Considering the model efficiency, we select QueryNet~\cite{chai2024querynet} as the IPS model, and SAM2.1 based on Hiera-B+~\cite{ryali2023hiera}.
% \DJI{take querynet as example model to validate our idea, actually we can apply our \ourmodel~to any IPS model. Can we conduct more experiments on extra IPS models, and supply them in the appendix?}\HQ{yes}For training phase, we train QueryNet independently, following their official setting, while for model inference, we formulate {\ourmodel} by combining QueryNet and SAM2 in a training-free manner.
% In terms of hyperparameters, we empirically set the length of time window $T$ to $3$ and the paired threshold $\theta$ to $0.5$ according to the ablation result in Table~\ref{table:4}, 
% and define $\lambda_1$ and $\lambda_2$ as $1$ and $3$, respectively, according to the results detailed in Appendix C.

% is implemented on the PyTorch framework~\cite{paszke2019pytorch} and using a single NVIDIA GeForce RTX 4090 GPU with $24$GB memory.
% Considering the model efficiency, we select Polyper~\cite{shao2024polyper} with PVT~\cite{wang2021pyramid} as the IPS model and SAM2 based on Hiera-B+~\cite{ryali2023hiera}.
% For training IPS models and fine-tuning SAM2, we follow the their provided raw training codes and settings.
% For model inference, we resize the input image into $352\times352$.
% In terms of hyperparameters, we set the time window $T$ and the association threshold $\tau$ to $3$ and $0.5$, respectively.

% \vspace{-5pt}
\subsection{In-domain Performance}
\label{sec:4.2}

\tabref{table:1} reports the ID performance of advanced polyp segmentation models, \ie, ``SUN-SEG (train set) $\Rightarrow$ SUN-SEG (test set)''. When trained on $\mathcal{D}_V$, {\ourmodel} outperforms its image-based counterpart, QueryNet, by $7.0\%$  in Dice and $4.1\%$ in TC. This improvement demonstrates that our method significantly improves the video segmentation performance based on existing IPS models. Furthermore, we outperform all other video-based models on the same training data, especially surpassing the runner-up, SALI \cite{hu2024sali}, by $1.9\%$ in Dice.
To further explore potential performance improvements, we investigate two strategies. First, we retrain the IPS model, QueryNet, using the combined set of $\mathcal{D}_V$ and $\mathcal{D}_I$. We observe that our \ourmodel~with QueryNet ($\mathcal{D}_V$+$\mathcal{D}_I$) achieved an IoU increase from $80.3\%$ to $82.5\%$, primarily due to enhanced spatial-context learning. Second, we unlock SAM2's spatiotemporal learning capacity by fully fine-tuning it on video data $\mathcal{D}_V$. 
Unless otherwise specified, \ourmodel~in the following analysis is built on QueryNet (trained on $\mathcal{D}_I$+$\mathcal{D}_V$), while SAM2 operates in a training-free manner, as indicated in the second-last row of \tabref{table:1}. 
In the upper part of \figref{fig3}, we qualitatively visualize the ID predictions of competing models in various settings. These visualizations demonstrate that {\ourmodel} effectively mitigates the discrete false-positive regions predicted by QueryNet, resulting in more accurate and temporally consistent segmentation results. Notably, when trained solely on $\mathcal{D}_V$, {\ourmodel} (fourth column) produces predictions that align more closely with ground-truth masks (last column) compared to the video model SALI (second column). Moreover, with the extra image data $\mathcal{D}_I$, the performance of QueryNet and {\ourmodel} is further refined.

\subsection{Out-of-domain Generalization}
% \subsection{Comparison in Trimmed Out-of-Domain Video Dataset}
\label{sec:4.3}

\begin{figure}[t!]
\centering
\includegraphics[width=0.95\textwidth]{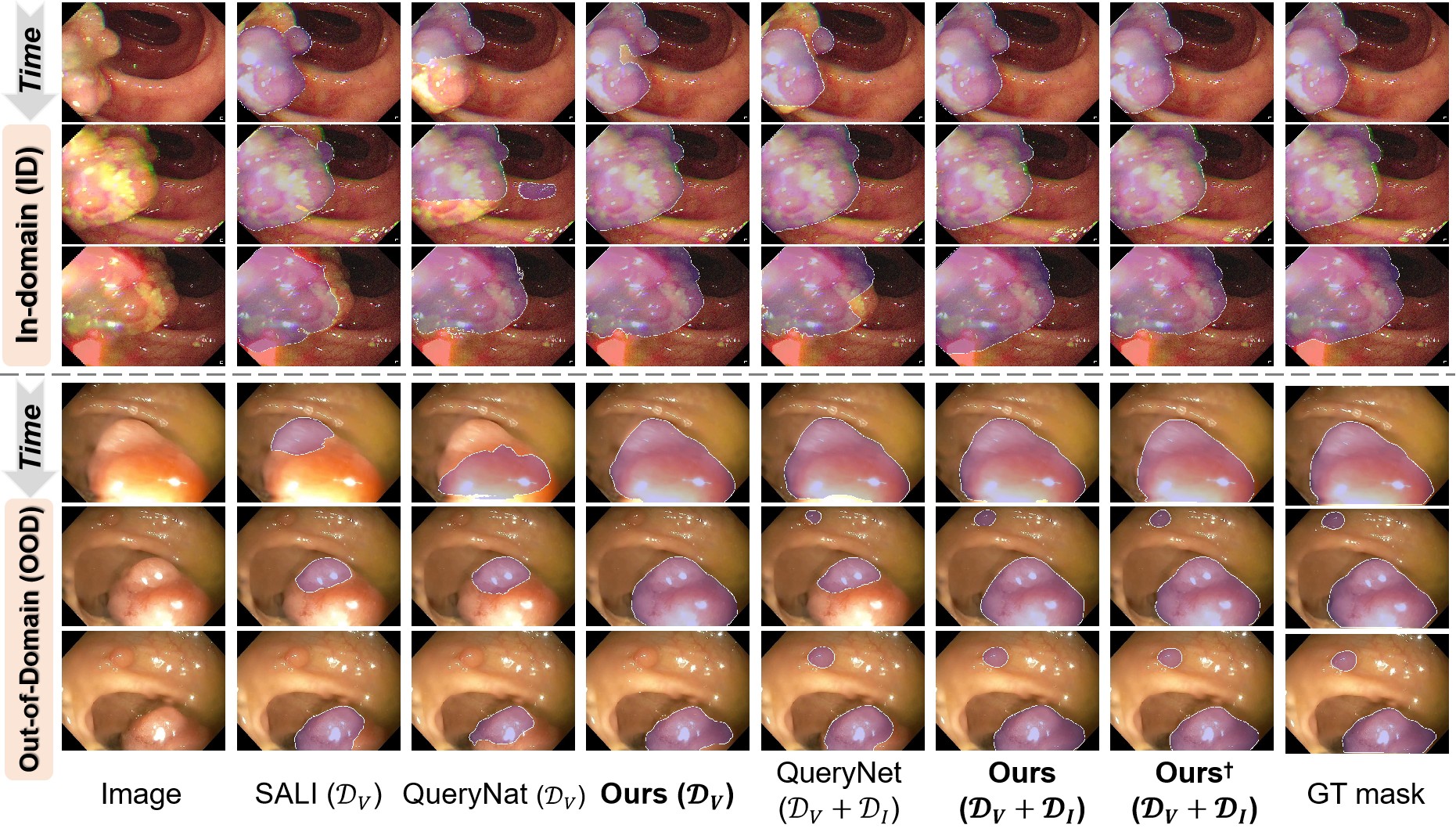}
\vspace{-5pt}
\caption{Prediction visualization on ID (SUN-SEG \cite{ji2022video}) and OOD (PolypGen \cite{ali2023multi}) datasets.}
\vspace{-10pt}
\label{fig3}
\end{figure}

Cross-domain generalization is a critical criterion for VPS models segmentation models in real clinical deployment.
To assess the OOD generalizability of all competing methods, we adopt two unseen VPS dataset to form a ``SUN-SEG (train set) $\Rightarrow$ PICCOLO/PolypGen'' evaluation pipeline.
As reported in \tabref{table:1}, when trained only on $\mathcal{D}_V$, {\ourmodel} outperforms other models on both unseen datasets, demonstrating robust cross-domain generalizability. 
With the inclusion of additional training data $\mathcal{D}_I$, both QueryNet and {\ourmodel} exhibit substantial improvements in OOD generalization.
Specifically, QueryNet improves Dice by $9.1\%$ and $12\%$  on PICCOLO and PolypGen, respectively, while {\ourmodel} achieves further gains of $4.2\%$ and $7.4\%$ in Dice.
This gain is primarily attributed to the diverse polyp appearances learned from $\mathcal{D}_I$, which provide richer contextual knowledge for representation.
As shown in the lower part of \figref{fig3}, training on the combined datasets (\ie, $\mathcal{D}_V$+$\mathcal{D}_I$) enables the model to identify previously missed, small-sized occult polyps, highlighting the benefits of enriched spatiotemporal learning.
% Moreover, {\ourmodel} demonstrates significant gains over its image counterpart, QueryNet, primarily due to its exploitation of the inherent zero-shot generalizability of SAM2. 
These results underscore the effectiveness of repurposing a robust video foundation model with our core innovations for enhanced ID-to-OOD generalization.

\subsection{Out-of-domain Generalization on Long-Untrimmed Colonoscopy Videos}\label{sec:4.4}

% \vspace{-10pt}

\begin{table*}
\begin{center}
\footnotesize
\caption{Comparisons on the in-house dataset ({\ourdata}). For segmentation models, the detection performance is evaluated by converting their output masks into the corresponding tightest bounding boxes. ``\textcolor{gray}{n/a}" denotes that the evaluation metric is not applicable.
% the results that we cannot measure, due to the detection model can not output pixel-level masks, and there is no confidence scores for the bounding box transformed from the semantic masks.
}
\vspace{-7pt}
\renewcommand\arraystretch{0.95}
\setlength{\tabcolsep}{5.2pt}
% \resizebox{0.95 \linewidth}{!}{
% \begin{small}
\begin{tabular}{lccc|ccc|cccc}
\hline

\hline
\rowcolor{mygray}
&&&&\multicolumn{3}{c|}{@Detection} &\multicolumn{2}{c}{@Segmentation}\\
\rowcolor{mygray}
Models &Model type &Train data &Label type &F1$_{\mathtt{50}}$ &AP$_{\mathtt{50}}$ &AP$_{\mathtt{50:95}}$ &Dice &IoU \\
\hline
\hline
% PraNet &$27.3$ &43.5 &33.5 &31.1 &25.6 &31.5  \\
YOLOv11~\cite{yolo11_ultralytics} &Image &$\mathcal{D}_{I}+\mathcal{D}_{V}$ &Box &$40.7$ &$37.1$ &$25.0$ &\textcolor{gray}{n/a} &\textcolor{gray}{n/a}\\
% IBoxCLA &35.3 &59.1 &44.2 &36.3 &29.0 &26.9\\
Relation-DETR~\cite{hou2024relation} &Image &$\mathcal{D}_{I}+\mathcal{D}_{V}$ &Box  &$44.0$ &$39.1$ &$26.7$ &\textcolor{gray}{n/a} &\textcolor{gray}{n/a}\\
TSdetector~\cite{wang2025tsdetector} &Video &$\mathcal{D}_{V}$ &Box  &$43.5$ &$42.2$ &$27.3$ &\textcolor{gray}{n/a} &\textcolor{gray}{n/a}\\
% Polyper &32.2 &63.0 &42.6 &$-$ &33.1 &26.5 &29.0 \\
% Polyper &37.4 &69.8 &48.7 &$-$ &41.2 &33.4 &22.8 \\
\hline
% PNS-Net &30.4 &59.6 &40.3 &$-$ &30.3 &23.6 &31.6 \\
% PNS+ &31.5 &61.0 &41.5 &$-$ &32.5 &24.1 &29.9 \\
Polyper~\cite{shao2024polyper} &Image &$\mathcal{D}_{I}+\mathcal{D}_{V}$ &Mask &$18.3$ &\textcolor{gray}{n/a} &\textcolor{gray}{n/a} &$58.7$ &$51.1$\\
Vivim~\cite{yang2024vivim} &Video &$\mathcal{D}_{V}$ &Mask &$14.6$ &\textcolor{gray}{n/a} &\textcolor{gray}{n/a} &$51.6$ &$43.5$\\
SALI~\cite{hu2024sali} &Video &$\mathcal{D}_{V}$ &Mask &$19.1$ &\textcolor{gray}{n/a} &\textcolor{gray}{n/a} &$55.0$ &$47.8$\\
\hline
 % &43.6 &68.6 &53.3 &39.8 &32.5 &24.0 \\
Mask2Former~\cite{cheng2022masked} &Image &$\mathcal{D}_{I}+\mathcal{D}_{V}$ &Box+Mask &$50.3$ &$47.7$ &$30.6$ &$61.8$ &$56.9$\\
QueryNet~\cite{chai2024querynet} &Image &$\mathcal{D}_{I}+\mathcal{D}_{V}$ &Box+Mask &$51.5$ &$45.6$ &$29.4$ &$63.7$ &$57.0$\\
\textbf{{\ourmodel} (Ours)} &Video &$\mathcal{D}_{I}+\mathcal{D}_{V}$ &Box+Mask &$\mathbf{63.0}$ &$\mathbf{52.4}$ &$\mathbf{33.8}$ &$\mathbf{68.2}$ &$\mathbf{62.8}$\\
% IAFR-VPS$^\dag$ &43.0 &64.1 &51.5 &40.1 &33.5 &24.8 \\
% IAFR-VPS$^\dag$ &\textbf{46.0} &\textbf{69.3} &\textbf{55.3} &\textbf{46.4} &\textbf{38.0} &\textbf{19.9} \\
\hline

\hline
\end{tabular}
% \end{small}
% }
\label{table:2}
\end{center}
\vspace{-0.7cm}
\end{table*}

Existing public VPS datasets usually presume that each frame contains at least one polyp, thus trimming videos into short clips of up to tens of seconds during data processing. However, in real clinical practice, raw colonoscopic videos often span tens of minutes, presenting more complex and challenging scenarios. To evaluate the long-term tracking capacity of competing models in clinical scenarios, we use a in-house dataset, \ourdata, to assess both temporal detection and segmentation performance over long-untrimmed videos. Specifically, we calculate the detection metrics (\ie, F1$_{\mathtt{50}}$, AP$_{\mathtt{50}}$, AP$_{\mathtt{50:95}}$) based on all frames and test the segmentation metrics (\ie, Dice, IoU) only on positive polyps. We include three categories of models for comparison: object detection methods \cite{yolo11_ultralytics, hou2024relation, wang2025tsdetector}, semantic segmentation methods \cite{shao2024polyper, yang2024vivim, hu2024sali}, and unified detection and segmentation methods \cite{cheng2022masked, chai2024querynet}. As shown in \tabref{table:2}, semantic segmentation models generally exhibit poorer detection performance compared to the other two types of models, which is attributed to their lack of architectural design and training objectives specifically for the detection task. In contrast, {\ourmodel} achieves the highest performance in both detection and segmentation tasks in long-untrimmed colonoscopy videos. This success underscores its potential for open-set clinical scenarios.

\subsection{Diagnostic Experiments}\label{sec:diagnostic_experiments}

\begin{wrapfigure}{r}{0.6\linewidth}
\vspace{-9pt}
\centering
\begin{overpic}
[width=0.6\textwidth]{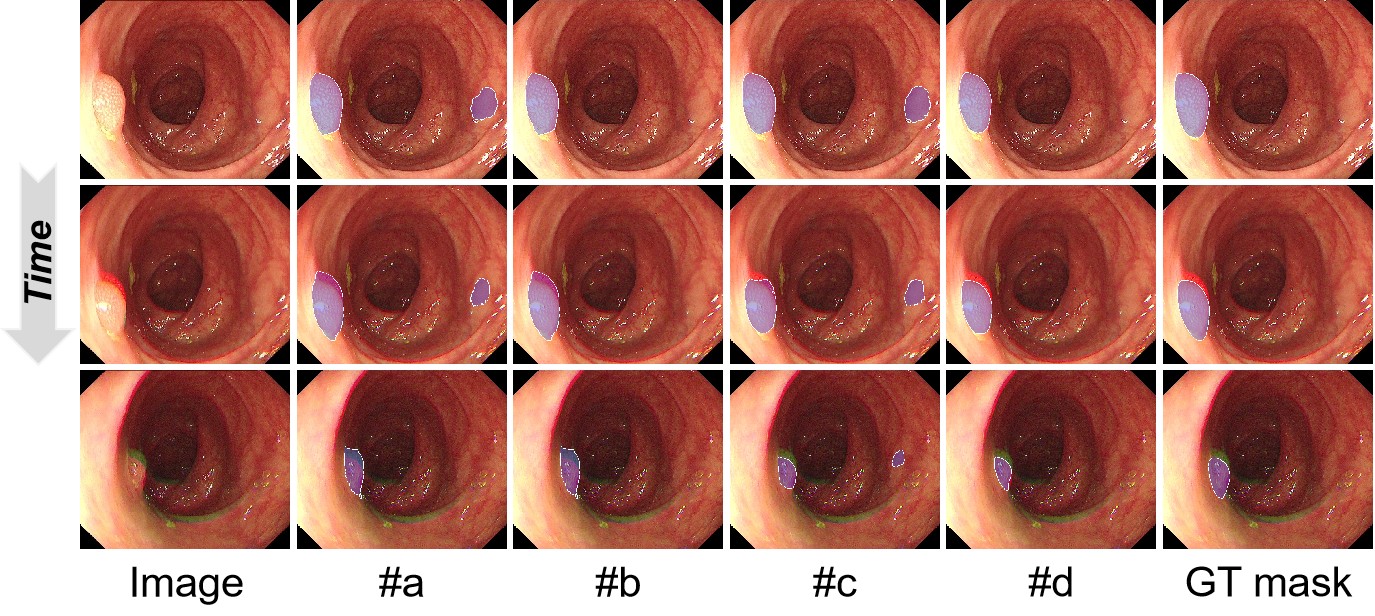}
\put(0,-13){
% \scriptsize
\setlength\tabcolsep{5.3pt}
\renewcommand\arraystretch{0.95}
\footnotesize
\begin{tabular}{c|r|cccc}
\hline

\hline
\rowcolor{mygray}
% &Dice &IoU &MAE &TC\\
% \rowcolor{mygray}
\rowcolor{mygray}
% IAF &IAR 
No. &Model variants & Dice &IoU &MAE &TC \\
\hline
\hline
% \multicolumn{2}{c|}{IPS Base.} & & 76.7& 68.0& 3.7\\
% \multicolumn{2}{c|}{SAM2 (GT)} & & & &\\
% \multicolumn{2}{c|}{SAM2 (IPS)} & & & &\\
% IPS Baseline &$82.2$ &$73.8$ &$2.8$ &$-$\\
% w/o IAR &$84.6$ &$76.1$ &$2.6$ &$-$\\
% w/o IAF &$81.2$ &$72.5$ &$3.5$ &$-$\\
% {\ourmodel} &$\mathbf{88.4}$ &$\mathbf{82.5}$ &$\mathbf{1.6}$ &$\mathbf{-}$\\
% \hline
% SAM2 (GT) &$75.5$ &$64.3$ &$4.6$ &$-$\\
% SAM2 (IPS) &$61.1$ &$53.7$ &$7.0$ &$-$\\
% \hline
\#a & QueryNet (baseline) &$82.2$ &$73.8$ &$2.8$ &$70.5$\\
\hline
\#b & + IAF &$84.6$ &$76.1$ &$2.6$ &$71.3$\\
\#c & + IAR &$81.2$ &$72.5$ &$3.5$ &$73.0$\\
\#d & \textbf{+ IAF \& IAR (Ours)} &$\mathbf{88.4}$ &$\mathbf{82.5}$ &$\mathbf{1.6}$ &$\mathbf{74.7}$\\
\hline

\hline
\end{tabular}
}
\end{overpic}
\captionsetup{width=0.6\textwidth}
\vspace{45pt}
\caption{Ablation study on IAF and IAR modules: qualitative (\textit{upper}) and quantitative results (\textit{lower}).}
\label{fig:4}
\vspace{-8pt}
\end{wrapfigure}

\paragraph{Effectiveness of Key Components.}
To validate the effectiveness of two key components of {\ourmodel}, the IAF and IAR modules, we test two variants of {\ourmodel} by disabling IAF or IAR, the IPS baseline QueryNet on SUN-SEG.
% \NB{something is broken in this sentence - not clear what IPS baseline refers to here} (\emph{i.e.}, QueryNet) 
% , and SAM2 directly driven by the ground truth mask and the IPS baseline prediction of the first frame on SUN-SEG.
% Note that if disabling IAF and IAR, the prediction results of the IPS model at each frame will be directly fed into the memory bank of SAM2 to drive it to segment the final result, which is taken as the baseline.
% As shown in Tab.~\ref{table:3}, when IAF is removed, the noise predictions of the IPS model mislead the entire framework, making the performance even lag behind the baseline in Dice, IoU and MAE, whereas the temporal modeling of method is lost when IAR is removed, compromising the temporal consistency.
As shown in the lower part of \figref{fig:4}, the model variant (\#c) illustrates that removing the IAF module causes the noise predictions from the IPS model to propagate throughout the pipeline, ultimately degrading performance and even falling behind the baseline in Dice, IoU, and MAE metrics. Similarly, as in model (\#b), removing the IAR module disrupts temporal modeling, affecting temporal consistency. 
More clearly in the upper part of \figref{fig:4}, the variant (\#c) retains the false positive prediction of the IPS model to be retained and be persistently segmented in the video stream.
The temporal stability of the segmentation of the variant (\#b) decreases, especially in the region of polyp boundaries.
In summary, IAF and IAR are mutually beneficial and together play an important role in combining the IPS model and SAM2.
% we can find that IAF and IAR are mutually beneficial and play an important role in combining the IPS model and SAM2, both compensating for the missing temporal consistency in the IPS model and providing continuous and stable driving information for SAM2.
% two key conclusions can be made as follows:
% (1) By comparing the first two rows in Table~\ref{table4}, if simply using IPS (Polyper) outputs to prompt SAM2, the baseline even lags behind the native IPS method, which demonstrates that simply combining the IPS method with SAM2 can lead to negative effects.
% (2) By comparing the last four rows in Table~\ref{table4}, we find that IAF and IAR improve the Dice of the baseline by $8.2\%$ and $4.7\%$, respectively, which shows the effectiveness of IAF and IAR.
% Next, by combining IAF and IAR, IAFR-VPS achieves the best performance, which shows that these two components are mutually beneficial.
% Notably, when IAF is excluded, the method still lags behind the native polyper, emphasizing the necessity of filtering the noisy predictions from the native IPS method.

% \tabref{table:3} (thrid row) illustrates that removing the IAF module causes the noise predictions from the IPS model to propagate throughout the pipeline, ultimately degrading the performance and even falling behind the baseline in Dice, IoU, and MAE metrics. Similarly, removing IAR disrupts temporal modeling, affecting temporal consistency. \figref{fig4} illusrates that without IAF

\begin{wraptable}{r}{0.58\linewidth}
\vspace{-12pt}
\centering
\footnotesize
\caption{Ablation study on IAF module.}
\vspace{-5pt}
\renewcommand\arraystretch{0.95}
\renewcommand\tabcolsep{8pt}
\begin{tabular}{r|ccc}
\hline

\hline
\rowcolor{mygray}
Model variants &Dice &IoU &MAE\\
\hline
\hline
\textit{w/o} align. &$84.1$ &$75.5$ &$2.5$\\
Optical flow-based align.~\cite{liu2020efficient} &$84.8$ &$76.5$ &$2.3$\\
% \rowcolor[rgb]{0.88,0.94,1}
\textbf{SAM2-based align. (Ours) }&$\mathbf{88.4}$ &$\mathbf{82.5}$ &$\mathbf{1.6}$\\
\hline
\textit{w/o} Voting filter (random) &$85.6$ &$78.2$ &$2.5$\\
% \rowcolor[rgb]{0.88,0.94,1}
\textbf{Voting filter (Ours)} &$\mathbf{88.4}$ &$\mathbf{82.5}$ &$\mathbf{1.6}$\\
\hline

\hline
\end{tabular}
\vspace{-8pt}
\label{table:3}
\end{wraptable}

\noindent\textbf{Effectiveness of the Sub-components in IAF.}
We conduct experiments to assess the effectiveness of cross-frame alignment and voting filter proposed in the IAF module. 
% For cross-frame alignment, we disable it and directly establish associations between IPS multiple predictions, and introduce an optical flow-based alignment method~\cite{liu2020efficient} for comparison.
For cross-frame alignment, we evaluate three settings: (1) disabling alignment and directly associating multiple IPS predictions, (2) applying an optical flow-based alignment method \cite{liu2020efficient}, and (3) our proposed SAM2-based alignment strategy.
% For the voting filter, we disable it and randomly select a segment from each tracklet.
Furthermore, we compare our voting filter method with a variant that randomly selects a segment from each tracklet.
% As can be seen from the results in \tabref{table:3}, both our proposed SAM2-based cross-frame alignment strategy and the voting filter strategy are superior.
% Optical flow-based alignment provides only a limited improvement over completely disabling alignment, mainly because of the very unstable lighting environment in endoscopic videos.
% Optical flow-based alignment provides only limited improvement, mainly due to the very unstable lighting environment in colonoscopy scenes.
As shown in Table~\ref{table:3}, both our SAM2-based alignment and voting filter are superior. Although optical flow-based alignment shows slight improvement over the no-alignment setting, its effectiveness is limited due to the unstable lighting conditions in the colonoscopy environment. In contrast, our proposed alignment method leverages SAM2's robust representations to achieve more consistent cross-frame associations, leading to substantial performance gains.

% The IAF module consists of two key sub-components: frame alignment and indicator voting.
% The former aims to mitigate the interference of frame shift on multi-frame association, while the latter ensures the selection of the best prediction from the associated frames.
% To verify the effectiveness of the two key sub-components in the IAF module, we conduct experiments on SUN-SEG.
% As shown in Table~\ref{table:4}, when frame alignment is disabled and IPS prediction associations across multiple frames are directly established, the average Dice of the four sub-test sets of SUN-SEG decreases by $4.5\%$.
% This demonstrates that frame shifts cause spatial bias in the predictions of the same polyp, making them difficult to associate correctly through simple IoU computation.
% Likewise, when indicator voting is eliminated and one of the associated segmentation results is randomly selected, the model’s segmentation performance drops by $1.7\%$ in terms of Dice, as this cannot guarantee the quality of the final result.

\begin{wraptable}{r}{0.6\linewidth}
\vspace{-13pt}
\centering
\footnotesize
\caption{Comparisons with other track-by-detect bridge methods on ID and OOD scenarios.}
\vspace{-5pt}
\renewcommand\arraystretch{1}
\renewcommand\tabcolsep{4.5pt}
\begin{tabular}{cl|cc|cc}
\hline

\hline
\rowcolor{mygray}
&&\multicolumn{2}{c|}{SUN-SEG} &\multicolumn{2}{c}{\ourdata}\\
\rowcolor{mygray}
&Methods &Dice &TC &Dice &TC\\
\hline
\hline
% DeepSORT~\cite{veeramani2018deepsort} \\
% \multirow{2}{*}{\begin{turn}{90}E2E\end{turn}} 
\multirow{2}{*}{End-to-End} &MaskTrack~\cite{yang2019video} &$82.8$ &$72.4$ &$50.4$ &$64.1$\\
&CTVIS~\cite{ying2023ctvis} &$85.7$ &$73.5$ &$58.1$ &$64.5$\\
\hline
% \multirow{3}{*}{\begin{turn}{90}Modular\end{turn}} 
\multirow{3}{*}{Modular}&DEVA~\cite{cheng2023tracking} &$79.8$ &$72.0$ &$62.6$ &$64.5$\\
&StrongSORT~\cite{du2023strongsort} &$83.0$ &$72.3$ &$64.3$ &$64.8$\\
% \rowcolor[rgb]{0.88,0.94,1}
&\textbf{\ourmodel~(Ours)} &$\mathbf{88.4}$ &$\mathbf{74.7}$ &$\mathbf{68.2}$ &$\mathbf{65.7}$\\
\hline

\hline
\end{tabular}
\vspace{-8pt}
\label{table:4}
\end{wraptable}

\noindent\textbf{Comparisons with Track-by-Detect Bridge Modules.}
We compare our {\ourmodel} with four track-by-detect bridge modules that can adapt image models to the video domain, including two end-to-end methods~\cite{yang2019video,cao2020sipmask} and two modular methods~\cite{cheng2023tracking,du2023strongsort}.
We uniformly set QueryNet as their image segmenters.
For end-to-end methods, we perform a two-stage training strategy, first training the image segmenter independently on the union of $\mathcal{D}_I$ and $\mathcal{D}_V$ and second training its entire framework on $\mathcal{D_V}$, while for modular methods, we implement them by directly combining the well-trained image segmenter on the union of $\mathcal{D}_I$ and $\mathcal{D}_V$ with their bridge modules.
Specially, we retrain a ReID model on polyp datasets for~\cite{du2023strongsort}.
% training the image segmenter independently on the union of $\mathcal{D}_I$ and $\mathcal{D}_V$ and then direct .
% we jointly train their entire framework only on $\mathcal{D_V}$, while for modular methods, we implement them by training the image segmenter independently on the union of $\mathcal{D}_I$ and $\mathcal{D}_V$.
% and implement modular-based methods by training image segmenters independently on $\mathcal{D}_I+\mathcal{D}_V$ and end-to-end methods by training their entire framework only on $\mathcal{D_V}$.
As shown in Tab.~\ref{table:4}, modular methods typically outperform end-to-end methods in OOD generalization because the bridge modules of end-to-end methods are trained only on limited polyp video data with limited generalization. Compared to motion matching and appearance matching used by~\cite{cheng2023tracking,du2023strongsort}, SAM2 implicitly learns more stable and unified spatio-temporal feature representations of objects from large-scale video data~\cite{ravi2024sam}, which contributes to the superiority of our method.

\begin{wraptable}{r}{0.6\linewidth}
\centering
\vspace{-12pt}
\caption{Ablation study on hyperparameters $T$ and $\theta$.}
\vspace{-6pt}
\renewcommand\arraystretch{0.95}
\renewcommand\tabcolsep{6.3pt}
\footnotesize
% \begin{small}
\begin{tabular}{cccc|cccc}
\hline

\hline
\rowcolor{mygray}
$T$ &Dice &IoU &MAE & $\theta$ &Dice &IoU &MAE\\
\hline
\hline
$2$ &$86.9$ &$80.8$ &$2.0$ &$0.3$ &$87.4$ &$81.7$ &$2.0$ \\
% \cellcolor[rgb]{0.88,0.94,1}$3$ &\cellcolor[rgb]{0.88,0.94,1}$\mathbf{88.4}$ &\cellcolor[rgb]{0.88,0.94,1}$\mathbf{82.5}$ &\cellcolor[rgb]{0.88,0.94,1}$\mathbf{1.6}$ &$0.4$ &$88.0$ &$81.9$ &$1.8$ \\
$3$ &$\mathbf{88.4}$ &$\mathbf{82.5}$ &$\mathbf{1.6}$ &$0.4$ &$88.0$ &$81.9$ &$1.8$ \\
% $4$ &$87.8$ &$81.5$ &$1.8$ &\cellcolor[rgb]{0.88,0.94,1}$0.5$ &\cellcolor[rgb]{0.88,0.94,1}$\mathbf{88.4}$ &\cellcolor[rgb]{0.88,0.94,1}$\mathbf{82.5}$ &\cellcolor[rgb]{0.88,0.94,1}$\mathbf{1.6}$ \\
$4$ &$87.8$ &$81.5$ &$1.8$ &$0.5$ &$\mathbf{88.4}$ &$\mathbf{82.5}$ &$\mathbf{1.6}$ \\
$5$ &$87.0$ &$80.3$ &$2.2$ &$0.6$ &$88.1$ &$82.1$ &$\mathbf{1.6}$ \\
\hline

\hline
\end{tabular}
% \end{small}
\label{table:5}
\end{wraptable}
\noindent\textbf{Hyperparameters configuration.}
% Here, we explore the setup of hyperparameters in our method.
% Tab.~\ref{table:6} presents the performance of {\ourmodel} on the SUN-SEG with different choices of hyperparameters: the length of time window $T$ and IAF association threshold $\theta$.
% For $T=2$, the voting filter represented in Eq.~(\ref{eq:4}) does not work, and we replace it with a randomly chosen one of the two segments from the two frames.
% We can see that {\ourmodel} with $T=3$ and $\theta=0.5$ shows the best performance.
% By default, we follow this definition of hyperparameters in {\ourmodel}.
As in \tabref{table:5}, we evaluate {\ourmodel}'s performance on SUN-SEG dataset with two hyperparameters: the length of time window $T$ and the association threshold $\theta$ of  IAF. With $T=2$, the voting filter in Eq.~(\ref{eq:4}) is ineffective, prompting us to select one of the two segments from the frames at random. We find that the best results occur with $T=3$ and $\theta=0.5$, which we use as our default configuration.
% Consequently, we default to these setups in {\ourmodel}.

%%%%%%%%%%%%%%%%%%%%%

% \definecolor{deepgreen}{rgb}{0, 0.5, 0}
\begin{wraptable}{r}{0.6\linewidth}
\vspace{-12pt}
\centering
\footnotesize
\caption{Versatility on YouTube-VIS 2019 \cite{yang2019video} \& 2021 \cite{YouTube-VIS-2021}.}
\vspace{-5pt}
\renewcommand\arraystretch{0.95}
\renewcommand\tabcolsep{3.75pt}
\begin{tabular}{r|ccc|ccc}
\hline

\hline
\rowcolor{mygray}
&\multicolumn{3}{c|}{YouTube-VIS-2019} &\multicolumn{3}{c}{YouTube-VIS-2021}\\
\rowcolor{mygray}
Models &AP &AP$_{50}$ &AR$_{1}$ &AP &AP$_{50}$ &AR$_{1}$\\
\hline
\hline
\multicolumn{7}{l}{$\blacktriangleright$~\emph{Image Segmenter: Mask R-CNN}~\cite{he2017mask}}\\
+ MaskTrack~\cite{yang2019video} &$30.3$ &$51.1$ &$31.0$ &$28.6$ &$48.9$ &$26.5$\\
\textbf{+ \ourmodel~(Ours)} &$\mathbf{42.3}$ &$\mathbf{66.0}$ &$\mathbf{41.9}$ &$\mathbf{40.2}$ &$\mathbf{62.0}$ &$\mathbf{35.3}$\\
\hline
\multicolumn{7}{l}{$\blacktriangleright$~\emph{Image Segmenter: CondInst}~\cite{tian2020conditional}}\\
+ CrossVIS~\cite{yang2021crossover} &$34.8$ &$54.6$ &$34.0$ &$33.3$ &$53.8$ &$30.1$\\
\textbf{+ \ourmodel~(Ours)} &$\mathbf{43.8}$ &$\mathbf{66.3}$ &$\mathbf{42.7}$ &$\mathbf{41.7}$ &$\mathbf{62.7}$ &$\mathbf{37.2}$\\
\hline

\hline
\end{tabular}
\label{table:6}
\end{wraptable}
% \subsection{Generality for other Image Segmentation Models}
% \begin{table*}
% \begin{center}
% \setlength{\tabcolsep}{5 pt}
% % \resizebox{0.95 \linewidth}{!}{
% \begin{small}
% \begin{tabular}{ll|ccc|ccc}
% \hline

% \hline
% \rowcolor{mygray}
% &&\multicolumn{3}{c|}{YouTube-VIS-2019} &\multicolumn{3}{c}{YouTube-VIS-2021}\\
% \rowcolor{mygray}
% \multirow{-2}{*}{Methods} &\multirow{-2}{*}{Backbone} &AP &AP$_{50}$ &AR$_{1}$ &AP &AP$_{50}$ &AR$_{1}$\\
% \hline
% \hline
% Mask R-CNN~\cite{he2017mask} + MaskTrack~\cite{yang2019video} &&$30.3$ &$51.1$ &$31.0$ &$28.6$ &$48.9$ &$26.5$\\
% Mask R-CNN + Ours &\multirow{-2}{*}{ResNet-50~\cite{he2016deep}} &&&\\
% \hline
% CondInst~\cite{tian2020conditional} + CrossVIS~\cite{yang2021crossover} &&$34.8$ &$54.6$ &$34.0$ &$33.3$ &$53.8$ &$30.1$\\
% CondInst + Ours &\multirow{-2}{*}{ResNet-50~\cite{he2016deep}} &&&\\
% \hline

% \hline
% \end{tabular}
% \end{small}
% % }
% \caption{Comparisons of various image segmentation models on video polyp segmentation tasks. `+ Ours' means transforming the image segmentation model into a video version by our way. All these image-based models are reproduced using their provided official codes.}
% \label{table:7}
% \end{center}
% \end{table*}
\noindent\textbf{Versatility for General Scenarios.}
{\ourmodel} is a plug-and-play framework that can be seamlessly integrated with any image segmentation model to enhance its performance in videos. We further validate {\ourmodel} versatility on general scenarios \cite{yang2019video,YouTube-VIS-2021} by adding it to two representative image segmenters \cite{he2017mask,tian2020conditional}. As reported in \tabref{table:6}, {\ourmodel} enables both image segmenters to achieve superior performance on video instance segmentation. Notably, unlike competing methods (\ie, MaskTrack~\cite{yang2019video}, CrossVIS~\cite{yang2021crossover}) that require training from scratch on video data, we directly use training-free SAM2 to achieve better performance, making it more deployment-friendly.

% combined with any image segmentation model to improve its performance in video segmentation tasks.
% To verify the generality of our method, we introduce two image segmentation models Mask R-CNN~\cite{he2017mask} and CondInst~\cite{tian2020conditional}.
% We report two track-by-detect methods~\cite{yang2019video,yang2021crossover} and our {\ourmodel} on YouTube-VIS-2019 $\mathtt{val}$ set and YouTube-VIS-2021 $\mathtt{val}$.
% As shown in Tab.~\ref{table:6}, our {\ourmodel} enables both models to achieve better performance on two video instance segmentation datasets.
% More importantly, unlike the other two competing methods which need to train from scratch on video data, {\ourmodel} can directly unite a pre-trained image segmenter without further training, which is more convenient.

\vspace{-0.2cm}
\section{Conclusion}
\vspace{-0.2cm}
In this work, we propose {\ourmodel}, which novelly implements the track-by-detect paradigm in VPS, effectively integrating the domain generalization of the IPS method with the spatiotemporal modeling of SAM2 in a training-free manner. To address SAM2's inherent limitation of snowballing errors, we propose two key cascade modules, IAF and IAR. Within each time window, IAR establishes associations between the IAR-filtered IPS predictions and SAM2 to dynamically and continuously manage the SAM2's memory bank, forming an automatic and efficient VPS framework.
Extensive experiments show that {\ourmodel} achieves SOTA performance on ID and OOD datasets compared to the existing IPS and VPS methods. In addition, {\ourmodel} shows promising performance in the in-house untrimmed colonoscopy video dataset. These results highlight the potential of {\ourmodel} for real clinical applications.
% {\ourmodel} can benefit from a better IPS model, but the existing IPS models remain underexplored.
% To address it, we aim to develop a robust polyp foundation model by collecting a large-scale and diverse polyp dataset in future work.
% Moreover, 
In future work, we plan to address the representation limitations of existing IPS models by developing a robust colonoscopy-specific foundation model. This will involve constructing a large-scale, diverse polyp dataset to enhance IPS capabilities and further strengthen the effectiveness of our {\ourmodel}.

\bibliographystyle{unsrt}
\bibliography{neurips_2025}

\appendix
\section{Public Datasets}
In this work, we mainly used nine public polyp segmentation datasets, including six image polyp segmentation (IPS) datasets: ClinicDB \cite{bernal2015wm}, Kvasir-SEG \cite{jha2020kvasir}, ColonDB \cite{tajbakhsh2015automated}, EndoScene \cite{vazquez2017benchmark}, ETIS \cite{silva2014toward} and PS-NBI2K \cite{yue2023benchmarking}, and three video polyp segmentation (VPS) datasets: SUN-SEG \cite{ji2022video}, PICCOLO \cite{sanchez2020piccolo}, PolypGen \cite{ali2023multi}.

\noindent{\textbf{Statistics.}}
We summarize the number of images/frames, videos, and official links for each public dataset in \tabref{appx:table:1}.

\begin{figure}[h]
\centering
\includegraphics[width=\textwidth]{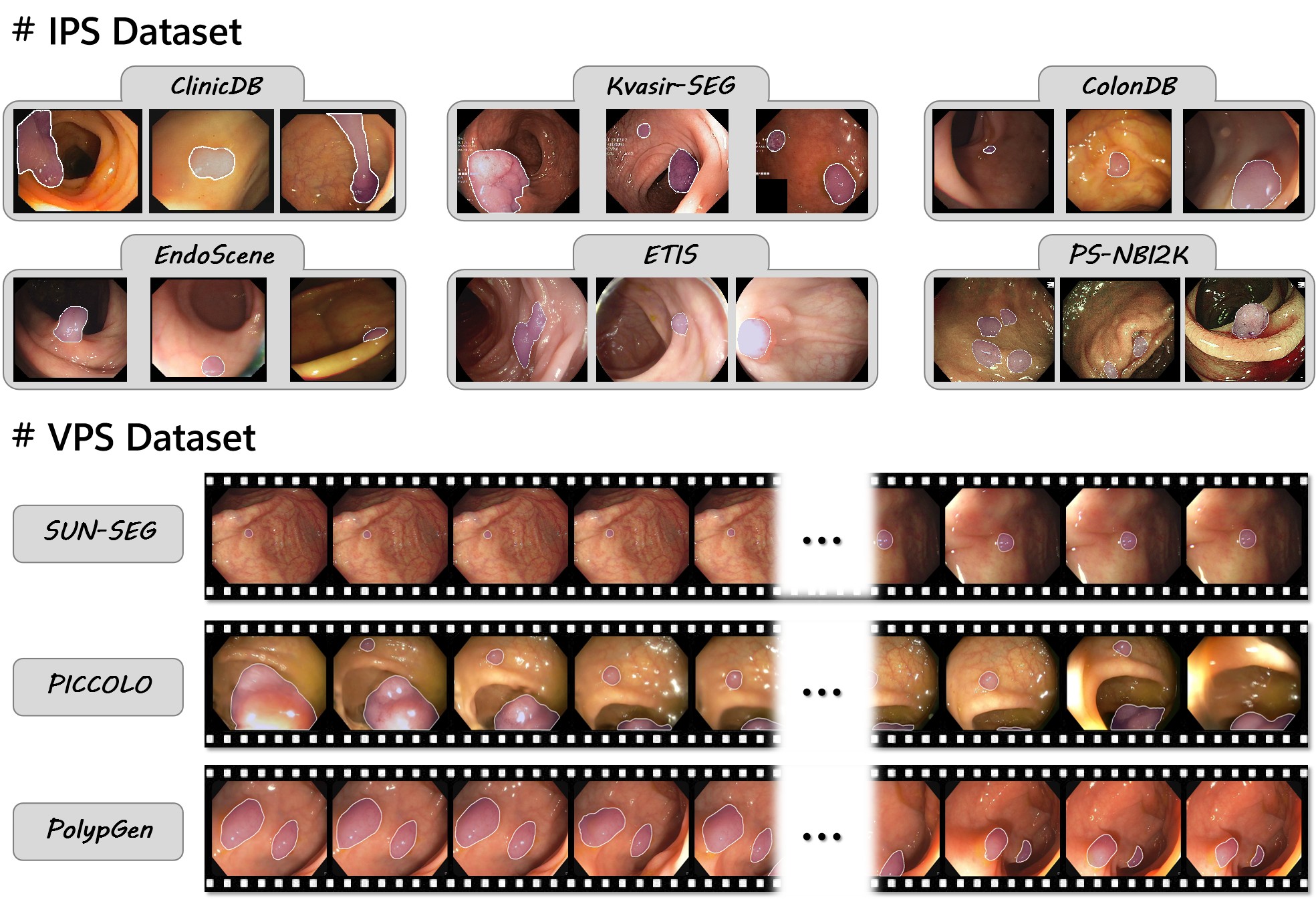}
\caption{Example cases from the public polyp segmentation datasets, including six IPS datasets and three VPS datasets.}
\label{appx:fig:dataset_visual}
\end{figure}

\noindent{\textbf{Visualization.}}
We select some example cases from each dataset and visualize them in \figref{appx:fig:dataset_visual}.
From it, we find that the IPS datasets contain diverse polyp types, which can enable the segmentation model to learn a more comprehensive polyp contextual knowledge, and thus improve the domain generalization of the polyp segmentation model.

\begin{table*}[h!]
\begin{center}
\caption{Statistics of public/in-house polyp segmentation datasets involved in our work. ``\#IMG" and ``\#VID" represents the number of images/frames and videos including in the dataset.}
\setlength{\tabcolsep}{17.5pt}
\renewcommand\arraystretch{1.2}
% \resizebox{0.95 \linewidth}{!}{
\begin{small}
\begin{tabular}{lllrrr}
\hline

\hline
\rowcolor{mygray}
Access &Type &Dataset &\#IMG &\#VID &URL\\
\hline
\hline
&& ClinicDB &612 &- &\href{https://polyp.grand-challenge.org/CVCClinicDB/}{Link}\\
&& Kvasir-SEG &1,000 &- &\href{https://datasets.simula.no/kvasir-seg/}{Link}\\
&& ColonDB &380 &- &\href{http://vi.cvc.uab.es/colon-qa/cvccolondb/}{Link}\\
&& EndoScene &60 &- &\href{https://pages.cvc.uab.es/CVC-Colon/index.php/databases/cvc-endoscenestill/}{Link}\\
&& ETIS &196 &- &\href{https://polyp.grand-challenge.org/ETISLarib/}{Link}\\
\multirow{-6}{*}{Public}&\multirow{-6}{*}{IPS}&PS-NBI2K &2,000 &- &\href{https://github.com/JaeZ1205/PS_NBI2k}{Link}\\
\hline
&& SUN-SEG (train) &19,544 &112 &\href{https://github.com/GewelsJI/VPS}{Link}\\
&& SUN-SEG (test) &29,592 &173 &\href{https://github.com/GewelsJI/VPS}{Link}\\
&& PICCOLO &3,433 &39 &\href{https://www.biobancovasco.org/en/Sample-and-data-catalog/Databases/PD178-PICCOLO-EN.html}{Link}\\
\multirow{-4}{*}{Public}&\multirow{-4}{*}{VPS}&PolypGen &8,037 &23 &\href{https://www.synapse.org/Synapse:syn26376615/wiki/613312}{Link}\\
\hline
In-house &VPS &LU-VPS &562,951 &15 &-\\

\hline

\hline
\end{tabular}
\end{small}
\label{appx:table:1}
\end{center}
\end{table*}
\vspace{-0.5cm}

\section{In-House Dataset: LU-VPS}
\begin{figure}[h]
\centering
\includegraphics[width=\textwidth]{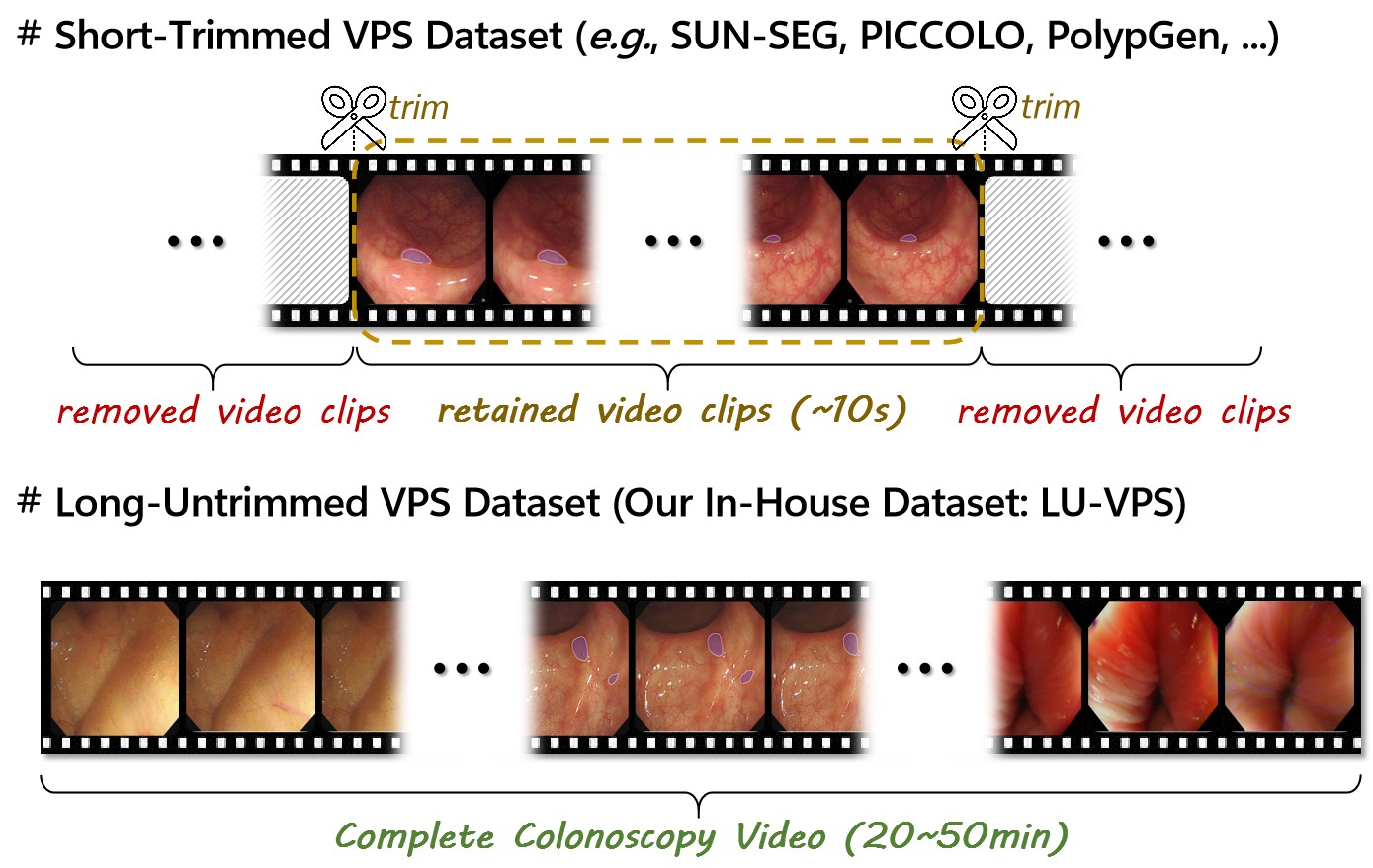}
\caption{Comparison between existing public short-trimmed VPS datasets and our in-house long-untrimmed VPS dataset.}
\label{appx:fig:trim_untrim}
\end{figure}

Existing VPS datasets including the three public video datasets mentioned above (\emph{i.e.,} SUN-SEG, PICCOLO, PolypGen) are all composed of \emph{short-trimmed} video clips, most of which are clips of polyps occurring consecutively.
The model's performance on these datasets can not reflect its performance in real clinical scenarios.
To address the limitation, we construct an in-house dataset, namely \textbf{{\ourdata}}, which contains a total of $9$ positive (containing polyps) and $6$ negative (without polyps) \emph{long-untrimmed} videos, totaling $562,951$ frames.
The length range of each video is $20\sim50$ minutes, with $15$ frames per second (FPS).
To more clearly illustrate the difference between the short-trimmed video clips and the long-untrimmed video clips, we show \figref{appx:fig:trim_untrim}.
Furthermore, we cropped a short clip from each video in the LU-VPS dataset and supplemented it in the path ``\textcolor{magenta}{\texttt{./Sample cases of LU-VPS (In-house Dataset)}}".

{\ourdata} was collected from a local hospital, with $59,104$ bounding box and $48,282$ pixel-wise mask annotations provided by three endoscopists to support the researchers in testing the detection and segmentation performance of the model.
Written informed consent was not required for this study as documented clinical colonoscopic images were collected retrospectively and appropriately by anonymizing and deidentifying.
Therefore, the study design was exempted from full review by the Institutional Review Board.

\section{Ablation on Two Hyperparameters: $\lambda_1$ and $\lambda_2$}
We conduct an ablation study on SUN-SEG dataset to examine the effect of the values of hyperparameters: $\lambda_1$ and $\lambda_1$, which are defined in Sec.~3.3 of manuscript.
$\lambda_1$ primarily affects the process by which newly discovered polyp instances are added to the SAM2's memory bank, whereas $\lambda_2$ primarily plays a role in the removal of information about disappearing polyps from the memory bank.

As reported in Tab.~\ref{appx:table:2}, we can see that \ourmodel~with $\lambda_1 = 1$ and $\lambda_2=3$ shows the best performance. Therefore, we follow this definition of hyperparameters in {\ourmodel} by default.
% We conduct the experiment on SUN-SEG dataset and report the results in Tab.~\ref{appx:table:2}. we can see that \ourmodel~with $\lambda_1 = 1$ and $\lambda_2=3$ shows the best performance. Therefore, we follow this definition of hyperparameters in \ourmodel~by default.
\begin{table*}[t]
\begin{center}
\caption{Ablation on two hyperparameters: $\lambda_1$ and $\lambda_2$.}
\setlength{\tabcolsep}{10.9 pt}
\renewcommand\arraystretch{1.2}
% \resizebox{0.95 \linewidth}{!}{
\begin{small}
\begin{tabular}{lcccc|lcccc}
\hline

\hline
\rowcolor{mygray}
$\lambda_1$ &Dice &IoU &MAE &TC &$\lambda_2$ &Dice &IoU &MAE &TC\\
\hline
\hline
$1$ &$\mathbf{88.4}$ &$\mathbf{82.5}$ &$1.6$ &$\mathbf{74.7}$ &$1$ &$83.2$ &$74.7$ &$2.7$ &$73.4$\\
% \cellcolor[rgb]{0.88,0.94,1}$3$ &\cellcolor[rgb]{0.88,0.94,1}$\mathbf{88.4}$ &\cellcolor[rgb]{0.88,0.94,1}$\mathbf{82.5}$ &\cellcolor[rgb]{0.88,0.94,1}$\mathbf{1.6}$ &$0.4$ &$88.0$ &$81.9$ &$1.8$ \\
$2$ &$88.0$ &$81.8$ &$\mathbf{1.5}$ &$74.5$ &$2$ &$87.4$ &$80.7$ &$1.8$ &$74.1$\\
% $4$ &$87.8$ &$81.5$ &$1.8$ &\cellcolor[rgb]{0.88,0.94,1}$0.5$ &\cellcolor[rgb]{0.88,0.94,1}$\mathbf{88.4}$ &\cellcolor[rgb]{0.88,0.94,1}$\mathbf{82.5}$ &\cellcolor[rgb]{0.88,0.94,1}$\mathbf{1.6}$ \\
$3$ &$87.5$ &$81.4$ &$1.8$ &$74.2$ &$3$ &$\mathbf{88.4}$ &$\mathbf{82.5}$ &$\mathbf{1.6}$ &$\mathbf{74.7}$ \\
$4$ &$87.2$ &$81.0$ &$1.8$ &$74.1$ &$4$ &$88.1$ &$82.0$ &$\mathbf{1.6}$ &$74.4$\\
\hline

\hline
\end{tabular}
\end{small}
\label{appx:table:2}
\end{center}
\end{table*}

\section{Generality for Other Image Polyp Segmentation Models}
{\ourmodel} is a plug-and-play framework and can improve the performance of any IPS model in VPS task.
To verify the generality of {\ourmodel}, we introduce IPS models, PolypPVT \cite{dong2021polyp} and Polyper \cite{shao2024polyper}.
Specifically, we retrain these IPS models using the combined set of $\mathcal{D}_V$ and $\mathcal{D}_I$, and then test their performance individually and when integrated with our {\ourmodel} on an in-domain (ID) dataset, SUN-SEG, and an out-of-domain (OOD) dataset, PICCOLO, respectively.

As shown in Tab.~\ref{appx:table:3}, we find that {\ourmodel} improves all metrics of all these IPS models on both ID and OOD datasets, which fully demonstrates the generality of our {\ourmodel} for various image polyp segmentation models.
\definecolor{deepgreen}{rgb}{0, 0.5, 0}
\begin{table*}[t]
\begin{center}
\caption{Verification of {\ourmodel 's} generality for other image polyp segmentation models.}
\setlength{\tabcolsep}{4.5 pt}
\renewcommand\arraystretch{1.2}
% \resizebox{0.95 \linewidth}{!}{
\begin{small}
\begin{tabular}{l|cccc|cccc}
\hline

\hline
\rowcolor{mygray}
& \multicolumn{4}{c|}{SUN-SEG (In Domain)} &\multicolumn{4}{c}{PICCOLO (Out-of-domain)}\\
\rowcolor{mygray}
{Methods} &Dice$\uparrow$ &IoU$~\uparrow$ &MAE~$\downarrow$ &TC~$\uparrow$ &Dice~$\uparrow$ &IoU~$\uparrow$ &MAE~$\downarrow$ &TC~$\uparrow$ \\
\hline
\hline
Polyp-PVT~\cite{dong2021polyp} &80.0 &71.9 &3.6 &69.8 &65.1 &58.3 &8.4 &65.5\\
\textbf{+ {\ourmodel}} &$86.1_{\textcolor{deepgreen}{{+6.1}}}$ &$79.6_{\textcolor{deepgreen}{{+7.7}}}$ &$2.6_{\textcolor{deepgreen}{{-1.0}}}$ &$73.7_{\textcolor{deepgreen}{{+3.9}}}$ &$73.0_{\textcolor{deepgreen}{{+7.9}}}$ &$66.4_{\textcolor{deepgreen}{{+8.1}}}$ &$6.5_{\textcolor{deepgreen}{{-1.9}}}$ &$68.9_{\textcolor{deepgreen}{{+3.4}}}$\\
\hline
Polyper~\cite{shao2024polyper} &$80.8$ &$73.0$ &$3.3$ &$70.3$ &$66.3$ &$59.1$ &$8.2$ &$65.8$ \\
\textbf{+ {\ourmodel}}  &$86.4_{\textcolor{deepgreen}{{+5.6}}}$ &$80.3_{\textcolor{deepgreen}{{+7.3}}}$ &$2.4_{\textcolor{deepgreen}{{-0.9}}}$ &$74.0_{\textcolor{deepgreen}{{+3.7}}}$ &$74.5_{\textcolor{deepgreen}{{+8.2}}}$ &$67.8_{\textcolor{deepgreen}{{+8.7}}}$ &$6.3_{\textcolor{deepgreen}{{-1.9}}}$ &$69.1_{\textcolor{deepgreen}{{+3.3}}}$ \\

% \midrule
% YOLOv11-seg~\cite{yolo11_ultralytics} &72.0 &66.2 &4.8 &61.5 &53.7 &8.2\\
% \textbf{+ {\ourmodel}} &72.0 &66.2 &4.8 &61.5 &53.7 &8.2\\
% \midrule
% Mask2Former~\cite{cheng2022masked} &82.5 &75.7 &2.6 &64.7 &58.9 &8.2\\
% \textbf{+ {\ourmodel}} &82.5 &75.7 &2.6 &64.7 &58.9 &8.2\\
% \hline

\hline
\end{tabular}
\end{small}
\label{appx:table:3}
\end{center}
\end{table*}

% \section{Generality for Nature Scene}
\section{More Visualization Results}
% \animategraphics[loop, controls, width=0.8\linewidth, autoplay, poster=0]{2}{appx_video1/video_v2-min_}{000}{009}
To provide a comprehensive visualization of {\ourmodel}'s performance, we have included supplementary GIFs in the Appendix zip.
These dynamic visualizations, which more intuitively demonstrates the superiority of {\ourmodel}, are organized in the path `` \textcolor{magenta}{\texttt{./Visualization results in ID and OOD scenarios}}".

\section{Limitations}
{\ourmodel} can adapt a pre-trained IPS model to the video domain in a training-free manner, although this has achieved a lot of success, there are still the following main limitations: (1) {\ourmodel} still relies on a reliable IPS model, which is a common problem in track-by-detect paradigm. We plan to collect a large-scale polyp dataset (including images and videos) to train a robust colonoscopy-specific foundation model, further enhancing the performance of {\ourmodel}. (2) The memory refinement mechanism will introduce latency in detecting newly appeared polyps, with a temporal delay of $0\!\sim\!T\cdot\lambda_1$ frames ($T=3$ and $\lambda_1=1$ in our default setting), even though this small delay does not affect the endoscopist's screening workflow.

\section{Broader Impacts}
{\ourmodel} achieves a better balance between spatiotemporal modeling and domain generalization compared to existing polyp segmentation methods and shows superior performance on the in-house long-untrimmed VPS dataset, which demonstrates the potential of {\ourmodel} in assisting clinical colonoscopy.
Further advancement in application requires subsequent validation in more rigorous clinical trials.
Moreover, {\ourmodel} eliminates the dependency on video-specific training data and is also expected to address the other medical video tasks where video data is scarce.

\newpage

\newpage
\section*{NeurIPS Paper Checklist}
\begin{enumerate}

\item {\bf Claims}
    \item[] Question: Do the main claims made in the abstract and introduction accurately reflect the paper's contributions and scope?
    \item[] Answer: \answerYes{} % Replace by \answerYes{}, \answerNo{}, or \answerNA{}.
    \item[] Justification: {Our abstract and introduction clearly state main contributions made in the paper. Our experimental results also support our claim.}
    \item[] Guidelines:
    \begin{itemize}
        \item The answer NA means that the abstract and introduction do not include the claims made in the paper.
        \item The abstract and/or introduction should clearly state the claims made, including the contributions made in the paper and important assumptions and limitations. A No or NA answer to this question will not be perceived well by the reviewers. 
        \item The claims made should match theoretical and experimental results, and reflect how much the results can be expected to generalize to other settings. 
        \item It is fine to include aspirational goals as motivation as long as it is clear that these goals are not attained by the paper. 
    \end{itemize}

\item {\bf Limitations}
    \item[] Question: Does the paper discuss the limitations of the work performed by the authors?
    \item[] Answer: \answerYes{}% Replace by \answerYes{}, \answerNo{}, or \answerNA{}.
    \item[] Justification: {We have a dedicated section to discuss the limitations of our work. Please refer to Appx. F.}
    \item[] Guidelines:
    \begin{itemize}
        \item The answer NA means that the paper has no limitation while the answer No means that the paper has limitations, but those are not discussed in the paper. 
        \item The authors are encouraged to create a separate "Limitations" section in their paper.
        \item The paper should point out any strong assumptions and how robust the results are to violations of these assumptions (e.g., independence assumptions, noiseless settings, model well-specification, asymptotic approximations only holding locally). The authors should reflect on how these assumptions might be violated in practice and what the implications would be.
        \item The authors should reflect on the scope of the claims made, e.g., if the approach was only tested on a few datasets or with a few runs. In general, empirical results often depend on implicit assumptions, which should be articulated.
        \item The authors should reflect on the factors that influence the performance of the approach. For example, a facial recognition algorithm may perform poorly when image resolution is low or images are taken in low lighting. Or a speech-to-text system might not be used reliably to provide closed captions for online lectures because it fails to handle technical jargon.
        \item The authors should discuss the computational efficiency of the proposed algorithms and how they scale with dataset size.
        \item If applicable, the authors should discuss possible limitations of their approach to address problems of privacy and fairness.
        \item While the authors might fear that complete honesty about limitations might be used by reviewers as grounds for rejection, a worse outcome might be that reviewers discover limitations that aren't acknowledged in the paper. The authors should use their best judgment and recognize that individual actions in favor of transparency play an important role in developing norms that preserve the integrity of the community. Reviewers will be specifically instructed to not penalize honesty concerning limitations.
    \end{itemize}

\item {\bf Theory assumptions and proofs}
    \item[] Question: For each theoretical result, does the paper provide the full set of assumptions and a complete (and correct) proof?
    \item[] Answer: \answerNA{} % Replace by \answerYes{}, \answerNo{}, or \answerNA{}.
    \item[] Justification: {The paper does not include theoretical results.}
    \item[] Guidelines:
    \begin{itemize}
        \item The answer NA means that the paper does not include theoretical results. 
        \item All the theorems, formulas, and proofs in the paper should be numbered and cross-referenced.
        \item All assumptions should be clearly stated or referenced in the statement of any theorems.
        \item The proofs can either appear in the main paper or the supplemental material, but if they appear in the supplemental material, the authors are encouraged to provide a short proof sketch to provide intuition. 
        \item Inversely, any informal proof provided in the core of the paper should be complemented by formal proofs provided in appendix or supplemental material.
        \item Theorems and Lemmas that the proof relies upon should be properly referenced. 
    \end{itemize}

    \item {\bf Experimental result reproducibility}
    \item[] Question: Does the paper fully disclose all the information needed to reproduce the main experimental results of the paper to the extent that it affects the main claims and/or conclusions of the paper (regardless of whether the code and data are provided or not)?
    \item[] Answer: \answerYes{} % Replace by \answerYes{}, \answerNo{}, or \answerNA{}.
    \item[] Justification: {The proposed model is described in Section 3. The detailed training setting and experimental setup are given in Section 4.2. All training datasets in this work are open-source and accessible. The trained model and code will be released after the review period.}
    \item[] Guidelines:
    \begin{itemize}
        \item The answer NA means that the paper does not include experiments.
        \item If the paper includes experiments, a No answer to this question will not be perceived well by the reviewers: Making the paper reproducible is important, regardless of whether the code and data are provided or not.
        \item If the contribution is a dataset and/or model, the authors should describe the steps taken to make their results reproducible or verifiable. 
        \item Depending on the contribution, reproducibility can be accomplished in various ways. For example, if the contribution is a novel architecture, describing the architecture fully might suffice, or if the contribution is a specific model and empirical evaluation, it may be necessary to either make it possible for others to replicate the model with the same dataset, or provide access to the model. In general. releasing code and data is often one good way to accomplish this, but reproducibility can also be provided via detailed instructions for how to replicate the results, access to a hosted model (e.g., in the case of a large language model), releasing of a model checkpoint, or other means that are appropriate to the research performed.
        \item While NeurIPS does not require releasing code, the conference does require all submissions to provide some reasonable avenue for reproducibility, which may depend on the nature of the contribution. For example
        \begin{enumerate}
            \item If the contribution is primarily a new algorithm, the paper should make it clear how to reproduce that algorithm.
            \item If the contribution is primarily a new model architecture, the paper should describe the architecture clearly and fully.
            \item If the contribution is a new model (e.g., a large language model), then there should either be a way to access this model for reproducing the results or a way to reproduce the model (e.g., with an open-source dataset or instructions for how to construct the dataset).
            \item We recognize that reproducibility may be tricky in some cases, in which case authors are welcome to describe the particular way they provide for reproducibility. In the case of closed-source models, it may be that access to the model is limited in some way (e.g., to registered users), but it should be possible for other researchers to have some path to reproducing or verifying the results.
        \end{enumerate}
    \end{itemize}

\item {\bf Open access to data and code}
    \item[] Question: Does the paper provide open access to the data and code, with sufficient instructions to faithfully reproduce the main experimental results, as described in supplemental material?
    \item[] Answer: \answerYes{} % Replace by \answerYes{}, \answerNo{}, or \answerNA{}.
    \item[] Justification: {All of the datasets involved in this work are open-source and accessible, except one in-house dataset for evaluating zero-shot generalization. The detailed descriptions of these datasets are presented in Section 4.1 and Appx.~B. The trained model and code will be released after the review period.}
    \item[] Guidelines:
    \begin{itemize}
        \item The answer NA means that paper does not include experiments requiring code.
        \item Please see the NeurIPS code and data submission guidelines (\url{https://nips.cc/public/guides/CodeSubmissionPolicy}) for more details.
        \item While we encourage the release of code and data, we understand that this might not be possible, so “No” is an acceptable answer. Papers cannot be rejected simply for not including code, unless this is central to the contribution (e.g., for a new open-source benchmark).
        \item The instructions should contain the exact command and environment needed to run to reproduce the results. See the NeurIPS code and data submission guidelines (\url{https://nips.cc/public/guides/CodeSubmissionPolicy}) for more details.
        \item The authors should provide instructions on data access and preparation, including how to access the raw data, preprocessed data, intermediate data, and generated data, etc.
        \item The authors should provide scripts to reproduce all experimental results for the new proposed method and baselines. If only a subset of experiments are reproducible, they should state which ones are omitted from the script and why.
        \item At submission time, to preserve anonymity, the authors should release anonymized versions (if applicable).
        \item Providing as much information as possible in supplemental material (appended to the paper) is recommended, but including URLs to data and code is permitted.
    \end{itemize}

\item {\bf Experimental setting/details}
    \item[] Question: Does the paper specify all the training and test details (e.g., data splits, hyperparameters, how they were chosen, type of optimizer, etc.) necessary to understand the results?
    \item[] Answer: \answerYes{} % Replace by \answerYes{}, \answerNo{}, or \answerNA{}.
    \item[] Justification: {The detailed experiment setting is described in Section 4.2. The trained model and code will be released after the review period.}
    \item[] Guidelines:
    \begin{itemize}
        \item The answer NA means that the paper does not include experiments.
        \item The experimental setting should be presented in the core of the paper to a level of detail that is necessary to appreciate the results and make sense of them.
        \item The full details can be provided either with the code, in appendix, or as supplemental material.
    \end{itemize}

\item {\bf Experiment statistical significance}
    \item[] Question: Does the paper report error bars suitably and correctly defined or other appropriate information about the statistical significance of the experiments?
    \item[] Answer: \answerNo{} % Replace by \answerYes{}, \answerNo{}, or \answerNA{}.
    \item[] Justification: {We follow the common practice in prior works and report the performance number on the standard benchmarks.}
    \item[] Guidelines:
    \begin{itemize}
        \item The answer NA means that the paper does not include experiments.
        \item The authors should answer "Yes" if the results are accompanied by error bars, confidence intervals, or statistical significance tests, at least for the experiments that support the main claims of the paper.
        \item The factors of variability that the error bars are capturing should be clearly stated (for example, train/test split, initialization, random drawing of some parameter, or overall run with given experimental conditions).
        \item The method for calculating the error bars should be explained (closed form formula, call to a library function, bootstrap, etc.)
        \item The assumptions made should be given (e.g., Normally distributed errors).
        \item It should be clear whether the error bar is the standard deviation or the standard error of the mean.
        \item It is OK to report 1-sigma error bars, but one should state it. The authors should preferably report a 2-sigma error bar than state that they have a 96\% CI, if the hypothesis of Normality of errors is not verified.
        \item For asymmetric distributions, the authors should be careful not to show in tables or figures symmetric error bars that would yield results that are out of range (e.g. negative error rates).
        \item If error bars are reported in tables or plots, The authors should explain in the text how they were calculated and reference the corresponding figures or tables in the text.
    \end{itemize}

\item {\bf Experiments compute resources}
    \item[] Question: For each experiment, does the paper provide sufficient information on the computer resources (type of compute workers, memory, time of execution) needed to reproduce the experiments?
    \item[] Answer: \answerYes{}{} % Replace by \answerYes{}, \answerNo{}, or \answerNA{}.
    \item[] Justification: {The information on computer resources is provided in Section 4.2.}
    \item[] Guidelines:
    \begin{itemize}
        \item The answer NA means that the paper does not include experiments.
        \item The paper should indicate the type of compute workers CPU or GPU, internal cluster, or cloud provider, including relevant memory and storage.
        \item The paper should provide the amount of compute required for each of the individual experimental runs as well as estimate the total compute. 
        \item The paper should disclose whether the full research project required more compute than the experiments reported in the paper (e.g., preliminary or failed experiments that didn't make it into the paper). 
    \end{itemize}
    
\item {\bf Code of ethics}
    \item[] Question: Does the research conducted in the paper conform, in every respect, with the NeurIPS Code of Ethics \url{https://neurips.cc/public/EthicsGuidelines}?
    \item[] Answer: \answerYes{} % Replace by \answerYes{}, \answerNo{}, or \answerNA{}.
    \item[] Justification: {Research conducted in the paper conforms with the NeurIPS Code of Ethics.}
    \item[] Guidelines:
    \begin{itemize}
        \item The answer NA means that the authors have not reviewed the NeurIPS Code of Ethics.
        \item If the authors answer No, they should explain the special circumstances that require a deviation from the Code of Ethics.
        \item The authors should make sure to preserve anonymity (e.g., if there is a special consideration due to laws or regulations in their jurisdiction).
    \end{itemize}

\item {\bf Broader impacts}
    \item[] Question: Does the paper discuss both potential positive societal impacts and negative societal impacts of the work performed?
    \item[] Answer: \answerYes{} % Replace by \answerYes{}, \answerNo{}, or \answerNA{}.
    \item[] Justification: {We have a dedicated section to discuss the societal impacts of our work. Please refer to Appx.~G.}
    \item[] Guidelines:
    \begin{itemize}
        \item The answer NA means that there is no societal impact of the work performed.
        \item If the authors answer NA or No, they should explain why their work has no societal impact or why the paper does not address societal impact.
        \item Examples of negative societal impacts include potential malicious or unintended uses (e.g., disinformation, generating fake profiles, surveillance), fairness considerations (e.g., deployment of technologies that could make decisions that unfairly impact specific groups), privacy considerations, and security considerations.
        \item The conference expects that many papers will be foundational research and not tied to particular applications, let alone deployments. However, if there is a direct path to any negative applications, the authors should point it out. For example, it is legitimate to point out that an improvement in the quality of generative models could be used to generate deepfakes for disinformation. On the other hand, it is not needed to point out that a generic algorithm for optimizing neural networks could enable people to train models that generate Deepfakes faster.
        \item The authors should consider possible harms that could arise when the technology is being used as intended and functioning correctly, harms that could arise when the technology is being used as intended but gives incorrect results, and harms following from (intentional or unintentional) misuse of the technology.
        \item If there are negative societal impacts, the authors could also discuss possible mitigation strategies (e.g., gated release of models, providing defenses in addition to attacks, mechanisms for monitoring misuse, mechanisms to monitor how a system learns from feedback over time, improving the efficiency and accessibility of ML).
    \end{itemize}
    
\item {\bf Safeguards}
    \item[] Question: Does the paper describe safeguards that have been put in place for responsible release of data or models that have a high risk for misuse (e.g., pretrained language models, image generators, or scraped datasets)?
    \item[] Answer: \answerNA{} % Replace by \answerYes{}, \answerNo{}, or \answerNA{}.
    \item[] Justification: {Our work does not pose such risks to the best of our knowledge.}
    \item[] Guidelines:
    \begin{itemize}
        \item The answer NA means that the paper poses no such risks.
        \item Released models that have a high risk for misuse or dual-use should be released with necessary safeguards to allow for controlled use of the model, for example by requiring that users adhere to usage guidelines or restrictions to access the model or implementing safety filters. 
        \item Datasets that have been scraped from the Internet could pose safety risks. The authors should describe how they avoided releasing unsafe images.
        \item We recognize that providing effective safeguards is challenging, and many papers do not require this, but we encourage authors to take this into account and make a best faith effort.
    \end{itemize}

\item {\bf Licenses for existing assets}
    \item[] Question: Are the creators or original owners of assets (e.g., code, data, models), used in the paper, properly credited and are the license and terms of use explicitly mentioned and properly respected?
    \item[] Answer: \answerYes{} % Replace by \answerYes{}, \answerNo{}, or \answerNA{}.
    \item[] Justification: {We properly credited the creators or original owners of assets (e.g., code, data, models), used in the paper and conformed the license and terms.}
    \item[] Guidelines:
    \begin{itemize}
        \item The answer NA means that the paper does not use existing assets.
        \item The authors should cite the original paper that produced the code package or dataset.
        \item The authors should state which version of the asset is used and, if possible, include a URL.
        \item The name of the license (e.g., CC-BY 4.0) should be included for each asset.
        \item For scraped data from a particular source (e.g., website), the copyright and terms of service of that source should be provided.
        \item If assets are released, the license, copyright information, and terms of use in the package should be provided. For popular datasets, \url{paperswithcode.com/datasets} has curated licenses for some datasets. Their licensing guide can help determine the license of a dataset.
        \item For existing datasets that are re-packaged, both the original license and the license of the derived asset (if it has changed) should be provided.
        \item If this information is not available online, the authors are encouraged to reach out to the asset's creators.
    \end{itemize}

\item {\bf New assets}
    \item[] Question: Are new assets introduced in the paper well documented and is the documentation provided alongside the assets?
    \item[] Answer: \answerYes{} % Replace by \answerYes{}, \answerNo{}, or \answerNA{}.
    \item[] Justification: {Our new assets introduced in the paper are well documented in the Section 4.1 and Appx.~B.}
    \item[] Guidelines:
    \begin{itemize}
        \item The answer NA means that the paper does not release new assets.
        \item Researchers should communicate the details of the dataset/code/model as part of their submissions via structured templates. This includes details about training, license, limitations, etc. 
        \item The paper should discuss whether and how consent was obtained from people whose asset is used.
        \item At submission time, remember to anonymize your assets (if applicable). You can either create an anonymized URL or include an anonymized zip file.
    \end{itemize}

\item {\bf Crowdsourcing and research with human subjects}
    \item[] Question: For crowdsourcing experiments and research with human subjects, does the paper include the full text of instructions given to participants and screenshots, if applicable, as well as details about compensation (if any)? 
    \item[] Answer: \answerNA{} % Replace by \answerYes{}, \answerNo{}, or \answerNA{}.
    \item[] Justification: {Our research does not involve crowdsourcing nor research with human subjects.}
    \item[] Guidelines:
    \begin{itemize}
        \item The answer NA means that the paper does not involve crowdsourcing nor research with human subjects.
        \item Including this information in the supplemental material is fine, but if the main contribution of the paper involves human subjects, then as much detail as possible should be included in the main paper. 
        \item According to the NeurIPS Code of Ethics, workers involved in data collection, curation, or other labor should be paid at least the minimum wage in the country of the data collector. 
    \end{itemize}

\item {\bf Institutional review board (IRB) approvals or equivalent for research with human subjects}
    \item[] Question: Does the paper describe potential risks incurred by study participants, whether such risks were disclosed to the subjects, and whether Institutional Review Board (IRB) approvals (or an equivalent approval/review based on the requirements of your country or institution) were obtained?
    \item[] Answer: \answerNA{} % Replace by \answerYes{}, \answerNo{}, or \answerNA{}.
    \item[] Justification: {Our research does not involve crowdsourcing nor research with human subjects.}
    \item[] Guidelines:
    \begin{itemize}
        \item The answer NA means that the paper does not involve crowdsourcing nor research with human subjects.
        \item Depending on the country in which research is conducted, IRB approval (or equivalent) may be required for any human subjects research. If you obtained IRB approval, you should clearly state this in the paper. 
        \item We recognize that the procedures for this may vary significantly between institutions and locations, and we expect authors to adhere to the NeurIPS Code of Ethics and the guidelines for their institution. 
        \item For initial submissions, do not include any information that would break anonymity (if applicable), such as the institution conducting the review.
    \end{itemize}

\item {\bf Declaration of LLM usage}
    \item[] Question: Does the paper describe the usage of LLMs if it is an important, original, or non-standard component of the core methods in this research? Note that if the LLM is used only for writing, editing, or formatting purposes and does not impact the core methodology, scientific rigorousness, or originality of the research, declaration is not required.
    %this research? 
    \item[] Answer: \answerNA{} % Replace by \answerYes{}, \answerNo{}, or \answerNA{}.
    \item[] Justification: {Our research does not involve LLMs as any important, original, or non-standard components.}
    \item[] Guidelines:
    \begin{itemize}
        \item The answer NA means that the core method development in this research does not involve LLMs as any important, original, or non-standard components.
        \item Please refer to our LLM policy (\url{https://neurips.cc/Conferences/2025/LLM}) for what should or should not be described.
    \end{itemize}

\end{enumerate}

\end{document}